\pdfoutput=1
\documentclass{article}

\usepackage[margin=1in]{geometry}
\usepackage{microtype}
\usepackage{graphicx}
\usepackage{subfigure}
\usepackage{hyperref}
\usepackage{amsmath,amssymb,amsthm}
\usepackage{mathtools}

\DeclareMathOperator{\sign}{sign}
\DeclareMathOperator*{\extr}{extr}
\DeclareMathOperator*{\argmin}{argmin}
\DeclareMathOperator*{\argmax}{argmax}
\DeclareMathOperator{\erf}{erf}

\theoremstyle{plain}
\newtheorem{result}{Result}[section]

\title{Asymptotic generalization error \\ of a single-layer graph convolutional network}

\author{\includegraphics[height=0.79em]{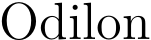} Duranthon and Lenka Zdeborová \\
 \small{Statistical Physics of Computation laboratory,}\\
 \small{École polytechnique fédérale de Lausanne (EPFL), Switzerland} \\
 \texttt{\small{firstname.lastname@epfl.ch}}}

\date{}

\begin{document}

\maketitle

\begin{abstract}
While graph convolutional networks show great practical promises, the theoretical understanding of their generalization properties as a function of the number of samples is still in its infancy compared to the more broadly studied case of supervised fully connected neural networks. 
In this article, we predict the performances of a single-layer graph convolutional network (GCN) trained on data produced by attributed stochastic block models (SBMs) in the high-dimensional limit. Previously, only ridge regression on contextual-SBM (CSBM) has been considered in \cite{shi2022statistical}; we generalize the analysis to arbitrary convex loss and regularization for the CSBM and add the analysis for another data model, the neural-prior SBM. We derive the optimal parameters of the GCN. We also study the high signal-to-noise ratio limit, detail the convergence rates of the GCN and show that, while consistent, it does not reach the Bayes-optimal rate for any of the considered cases.
\end{abstract}

\section{Introduction and related work}
Understanding the generalization properties of neural networks on unseen data is still unsatisfactory despite the very active line of work in this direction. In this article, we are specifically interested in understanding the generalization properties of graph neural networks, where the question remains even further from closed compared to feedforward neural networks that have been explored more broadly in the theoretical literature. 

\paragraph{Tight analysis in the high-dimensional limit:}
The question of generalization has been studied from many angles. Classical learning theory usually aims to avoid assumptions on the data distribution and to provide generic generalization bounds. Such bounds are, however, often far away from the actual performance on given benchmark datasets, see e.g. \cite{zhang2021understanding}. This generic line of work is hence complemented by studies of concrete data distributions and concrete target functions. Tight theoretical results are attainable in the high-dimensional limit, where the number of samples and their dimension go to infinity while being proportional. In this limit many quantities of interest concentrate on deterministic values for which a closed-set of dimension-independent fixed point equations is derived; see e.g. \cite{gardner1989three,barbier2019optimal,mei2022generalization,montanari2019generalization,aubin20glmReg}. This nice property is referred to as the blessing of dimensionality. 
This line of theoretical analysis is very appealing because it is able to provide results for the information-theoretically attainable generalization error, as well as the one obtained by a specific neural network. This allows us to evaluate the gap between the generalization ability of neural networks and the information-theoretically optimal one. The amplitude of the gaps to optimality can then be used to drive the development of architectures and algorithms that decrease the gap. The behaviour of systems of moderate sizes converges very fast to the asymptotic behaviour derived in the high-dimensional limit, thus making it relevant and interesting, as shown by the above works.
The main drawback of this line of work is that so far the available theoretical tools only allow such analysis for only very simple network architectures, e.g. single layer and two-layer neural networks. Still, there are many open questions for the two-layer case \cite{Cui_extensive_width}; and even for the simpler single-layer case, which corresponds to high-dimensional regressions, many open questions have been settled only recently: see e.g. \cite{sur2019modern,hastie2022surprises,aubin20glmReg}. However, the long-term promise of this direction of research motivates efforts to establish the tight asymptotic analysis and the underlying tools in broader and broader settings. The present work is inscribed in this context and it treats a graph convolutional neural network. In the same sense as done in the literature for the feed-forward fully connected networks, we will consider only single-layer graph convolutional networks (GCNs). This is a clear limitation of our work which is justified by the technical challenge of that setting and by the overall aim to build theoretical tools and understanding that will be able to deal with more realistic architectures in the future. Yet, on a practical point of view, linear single-layer GCNs can have similar performances to non-linear multi-layer ones, while being able to deal with very large graphs and being much simpler to train, as shown by \cite{wu19simpleGCN} and \cite{zhu21simpleGCN}.

\paragraph{Generalization in graph neural networks:}
Graph neural networks (GNNs) show a broad range of practical applications, and, as such, understanding their generalization properties is an important part of our overall goal. Many works consider graph or node classification in a learning scenario where one has access to many training graphs and unseen test graphs. Some works then derive bounds based on VC dimension, Rademacher complexity or PAC-Bayesian analysis, see for instance \cite{ju23pacBayes} and the references therein; wide networks can be analyzed thanks to graph neural tangent kernel, see e.g. \cite{qin23gntk}.
Instead, we consider the semi-supervised (or transductive) learning scenario, where training and inference are done on the same large graph whose node labels are partially revealed. This setting is relevant for node classification problems such as community detection. Previous theoretical works on semi-supervised learning include \cite{tang23generalizationGNN}, which studies learning under stochastic gradient descent, or \cite{cong21depthGCN} that focuses on graph convolutional networks and proposes experiments on data generated by the contextual stochastic block model (CSBM). More similar questions to our work are addressed in \cite{esser21generalizationCSBM} that derives generalization bounds for a particular model of data close to the CSBM, yet considering a generic GNN. These three works derive only loose bounds for the test performances of the GNN and they do not provide insights on the effect of the structure of data, such as its heterophily. For instance \cite{esser21generalizationCSBM} derives bounds based on transductive Rademacher complexity; since they are too general the authors have to model the data as a CSBM. Still the error bound they obtain is increasing with the number of samples $N$, which in the limit of large $N$ provides no guarantee. \cite{tang23generalizationGNN} provides sharper bounds; yet they are not tight, do not take in account the data and depend on continuity constants that cannot be determined a priori. A series of works closer to our article has been developed by the authors of \cite{fountoulakis21GC}. In this work, they consider a one-layer GCN trained on the CSBM by logistic regression and derive bounds for the test loss; however, they analyze its generalization ability on new graphs that are independent of the train graph and do not give exact predictions. In \cite{baranwal23clipGNN} they propose an architecture of GNN that is optimal for the CSBM, among classifiers that process local tree-like neighborhoods, and they exactly derive its generalization error. These two works consider a low-dimensional setting.

The tight analysis of generalization in synthetic high-dimensional settings for GNNs is still in its infancy. The only pioneering reference in this direction we are aware of is \cite{shi2022statistical} where the authors consider a simple one-layer GCN trained in a semi-supervised way by ridge regression. They predict its asymptotic performances on data generated by the CSBM and, in particular, show how to tune the architecture to adapt to the homophily strength of the graph.

A starting point of the tight asymptotic analysis of generalization is a suitable model for generating data. As \cite{shi2022statistical} showed, the CSBM introduced in \cite{yan2021covariate,cSBM18} is suitable. Data generated by this model has been used to benchmark various GNN architectures in \cite{pageRankGNN20,cong21depthGCN,pLaplacianGNN21,evenNet22} for instance. Another way to generate graph data with node features is the neural-prior or generalized linear model SBM (GLM--SBM) introduced in \cite{dz23glmSbm}, where the features alone do not bring any information. For these two models, the CSBM and the GLM--SBM, the optimal performance has been derived in the high-dimensional limit in \cite{dz23csbm,aPrioriGen19,dz23glmSbm}.

Our motivation to extend \cite{shi2022statistical} comes from the related line of research we detailed above. This work does not compare the performance of the GCN to the Bayes-optimality nor study the interplay between the loss, the regularization and the data; while, as to high-dimensional regression, \cite{aubin20glmReg,mignacco20gaussMixt} established that the generalization error of the ridge regression is suboptimal for some models of data while logistic regression is much closer to the Bayes-optimality. When it comes to rates with which the test error goes to zero in the limit of a large number of samples, they are again suboptimal for ridge regression while they give the Bayes-optimal rates for optimally regularized logistic regression \cite{aubin20glmReg}. For a slightly different setting the Bayes-optimal performance can be achieved \cite{mignacco20gaussMixt} just by adjusting the regularization. Natural questions thus are: how does the performance of the GCN from \cite{shi2022statistical} compare to the Bayes-optimal performance? How much do optimal regularization, architecture or loss improve the generalization? How does this reflect in rates when the signal-to-noise ratio is large? These questions are answered in the present article.

\paragraph{Main contribution:}
First we generalize the analysis of \cite{shi2022statistical} by considering generic loss and regularization for the CSBM and the GLM--SBM. We derive the summary statistics and the self-consistent equations they follow, which allow us to predict the exact generalization performance of the GCN in the high-dimensional limit. We show that these predictions are in very good agreement with numerical simulations of the GCN at finite $N$.

Using these predictions we compare to the Bayes-optimal test accuracy, search for the optimal parameters of the considered architecture and explore several common loss functions. We show that in the considered setting large regularization maximizes the test accuracy for the CSBM while leading to a test accuracy close to the optimum for the GLM--SBM; ridge regression has a large gap to the optimality, and the logistic and hinge losses do not improve it significantly. This stands for both the considered models and is thus different from the single-layer perceptron learning from data generated by the teacher-student model of \cite{aubin20glmReg}.
We derive an explicit formula for the test accuracy in the limit of large regularization, that allows us to make further predictions and understand rather explicitly the trade-off between how the GCN uses the graph and the features.
Then we take the limit of high signal-to-noise ratio (snr). We show that the simple GCN we consider is consistent in the sense that the test error converges to zero as the snr diverges. We derive the convergence rates for the two models; they appear to be smaller than the Bayes-optimal one, which is again in disparity with the well-studied feed-forward case \cite{aubin20glmReg}.
Last we derive the optimal self-loop strength of the GCN and provide evidence that this prediction may be generalizable to a broader class of datasets.

\section{Models, setup}
\label{sec:modèle}
\paragraph{Attributed SBMs:}
We consider a set of $N$ nodes and a graph $G$. Each node $i$ has a label $y_i=\pm 1$; we consider two balanced groups. We precise the law of $y_i$ later. We observe an adjacency matrix $A\in\mathbb R^{N\times N}$ whose components are drawn according to
\begin{equation}
A_{ij} \sim \mathcal B\left(\frac{d}{N}+\frac{\lambda}{\sqrt N}\sqrt{\frac{d}{N}\left(1-\frac{d}{N}\right)}y_iy_j\right)
\end{equation}
where $\lambda$ is the signal-to-noise ratio (snr) of the graph, $d$ is the average degree of the graph, $\mathcal B$ is a Bernoulli law and the components $A_{ij}$ are independent random variables. We take an average degree $d$ of order $N$, but $d$ growing with $N$ should be sufficient for our results to hold. We discuss this assumption more in detail in the appendix~\ref{sec:équGauss}. We consider a directed SBM, $A$ non-symmetric, to simplify the analysis; yet this model can be mapped to a non-directed SBM of snr $\lambda'=\sqrt 2\lambda$ by taking the adjacency matrix $(A+A^T)/\sqrt 2$.

We consider $M$ hidden independent standard Gaussian variables $u_\nu$; we set $\alpha=N/M$ the aspect ratio. We also observe features $X\in\mathbb R^{N\times M}$. The features are correlated with the node labels. We consider first the contextual stochastic block model (CSBM) \cite{yan2021covariate,cSBM18} for which the labels are Rademacher and the features follow a Gaussian mixture:
\begin{equation}
(\mathrm{CSBM})\quad y_i\sim\mathrm{Rad}\ , \; X = \sqrt\frac{\mu}{N}yu^T+W \label{eq:csbm}
\end{equation}
where $\mu$ is the snr of the features and $W$ is noise whose components $W_{i\nu}$ are independent standard Gaussians.
We will also consider another related model, the neural-prior or GLM--SBM \cite{dz23glmSbm}, for which the features are Gaussian and the labels are generated by a generalized linear model (GLM) on the features, the $\sign$ being applied element-wise:
\begin{equation}
(\mathrm{GLM-SBM})\quad X_{i\nu}\sim\mathcal N(0,1)\ , \; y=\sign\left(\frac{1}{\sqrt N}Xu\right)\ . \label{eq:glmSbm}
\end{equation}

We are given a set $R$ of train nodes and define $\rho=|R|/N$ the training ratio. The test set $R'$ is selected from the complement of $R$; we define $\rho'=|R'|/N$ as the testing ratio. We assume that $R$ and $R'$ are independent from the other quantities. The inference problem is to find back $y$ and $u$ given $A$, $X$, $R$ and the parameters of the model.

We work in the high-dimensional limit $N\to\infty$ and $M\to\infty$ while the aspect ratio $\alpha=N/M$ is of order one. The other parameters $\lambda$, $\mu$, $\rho$ and $\rho'$ are also of order one.

We precise that the total snr of the symmetric CSBM and GLM--SBM are \cite{cSBM18,dz23glmSbm}
\begin{equation}
\mathrm{snr}_\mathrm{CSBM}=\lambda^2+\frac{\mu^2}{\alpha}\ , \qquad \mathrm{snr}_\mathrm{GLM-SBM}=\lambda^2\left(1+\frac{4\alpha}{\pi^2}\right)\ . \label{eq:snrCsbm-GlmSbm}
\end{equation}
Authors of \cite{cSBM18,cSBM20,dz23glmSbm} established that $\mathrm{snr}_\mathrm{CSBM}=1$ and $\mathrm{snr}_\mathrm{GLM-SBM}=1$ are the detectability thresholds in the sense that in the unsupervised case $\rho=0$ they separate an undetectable phase, where the labels $y$ cannot be recovered better than at random, from a detectable phase where they can. In the semi-supervised case $\rho>0$ this transition disappears and one can always recover some information on the test labels.
The expression of $\mathrm{snr}_\mathrm{CSBM}$ shows that the snr originating from the graph is of the strength $\lambda^2$ while the one originating from the features is $\mu^2/\alpha$.

\paragraph{Analyzed GCN architecture:}
We follow \cite{shi2022statistical} and we consider a single-layer graph convolutional network (GCN). It transforms the features according to
\begin{equation}
h(w) = \frac{1}{N}Q(\tilde A)Xw
\end{equation}
where $Q$ is a polynomial, $w\in\mathbb R^M$ are the trainable weights and $\tilde A\in\mathbb R^{N\times N}$ is a rescaling of the adjacency matrix defined by
$\tilde A_{ij}=\left(\frac{d}{N}\left(1-\frac{d}{N}\right)\right)^{-1/2}\left(A_{ij}-\frac{d}{N}\right)$.
For the analysis, we consider $Q$ of degree one as in \cite{shi2022statistical}, i.e. $Q(\tilde A)=\tilde A+c\sqrt NI_N$ where $c$ is a tunable parameter of the architecture. This corresponds to applying one step of graph convolution to the features with self-loops.

This GCN is trained by empirical risk minimization. We define the regularized loss
\begin{equation}
L_{A,X}(w) = \frac{1}{\rho N}\sum_{i\in R}l(y_ih_i(w))+\frac{r}{\rho N}\sum_\nu \gamma(w_\nu)
\label{eq:loss}
\end{equation}
where $\gamma$ is a strictly convex regularization function, $r$ is the regularization strength and $l$ is a convex loss function. We will focus on $l_2$-regularization $\gamma(x)=x^2/2$ and on the square loss $l(x)=(1-x)^2/2$, the logistic loss $l(x)=\log(1+e^{-x})$ or the hinge loss $l(x)=\max(0,1-x)$. Since $L$ is strictly convex it admits a unique minimizer $w^*$. The average train and test errors and accuracies of this model are
\begin{align}
E_\mathrm{train/test} = \mathbb E\ \frac{1}{|\hat R|}\sum_{i\in\hat R}l(y_ih(w^*)_i)\ , \quad\quad
\mathrm{Acc}_\mathrm{train/test} = \mathbb E\ \frac{1}{|\hat R|}\sum_{i\in\hat R\ ,}\delta_{y_i=\sign{h(w^*)_i}}
\end{align}
where $\hat R$ stands either for the train set $R$ or the test set $R'$ and the expectation is taken over $y$, $u$, $A$, $X$, $R$ and $R'$.
We want to stress our reasons behind the choice of such a simple architecture. As discussed in the introduction, even for the more widely studied feed-forward fully connected neural networks, the generalization properties from a limited amount of training data is only properly understood in the single-layer case and partly for two-layer neural networks. A tight analysis for these cases is already challenging and actively developed. We extend this line to GCNs, which is a non-trivial task. The long-term goal is to build analysis tools and techniques to be able to tackle more complete architectures. Doing that directly is beyond the reach of the current theoretical toolbox.  
{\small
\begin{table}[h!]
\caption{Summary of the parameters of the model.}
\begin{tabular}{cl}
$N$ & size of the graph \\
$M$ & dimensionality of the attributes \\
$\alpha=N/M$ & aspect ratio \\
$d$ & average degree of the graph \\
$\lambda$ & snr of the SBM \\
\end{tabular}
\begin{tabular}{cl}
$\mu$ & snr of the Gaussian mixture \\
$l$, $\gamma$ & loss and regularization functions \\
$\rho = |R|/N$ & fraction of training nodes \\
$r$ & regularization strength \\
$c$ & self-loop strength
\end{tabular}
\end{table}
}
\paragraph{Bayes-optimal performances:}
An important consequence of modeling the data as we propose is that one has access to the Bayes-optimal (BO) performance on this task, i.e. the upper-bound on the test accuracy that any algorithm can reach, knowing the model and its parameters $\alpha, d, \lambda, \mu$. It is of particular interest since it will allow us to check how far the GCN is from the optimality and how much improvement can be done. The BO performances for both the CSBM and the GLM--SBM have been derived in \cite{dz23csbm,aPrioriGen19,dz23glmSbm}. They can be expressed as a function of the solution of the equations reproduced in appendix \ref{sec:eqBO}.

\section{Asymptotic prediction of the performances of the GCN}
\label{sec:theory}
In this section we state our main result, namely the asymptotic formulae for the expected losses and accuracies of the trained GCN. We will derive several consequences from these in the next section. We introduce the order parameters of the model and give the fixed-point equations they satisfy. We express the expected losses and accuracies as a function of these.

\begin{result}[Performances on the CSBM]
We consider the high-dimensional limit defined in the previous section.
Let $u$, $\varsigma$, $\xi$, $\zeta$ and $\chi$ be standard Gaussian random variables and $y$ be a Rademacher random variable. Let $\Theta = \{m_w, m_\sigma, Q_w, Q_\sigma, V_w, V_\sigma\}$ and $\hat\Theta = \{\hat m_w, \hat m_\sigma, \hat Q_w, \hat Q_\sigma, \hat V_w, \hat V_\sigma\}$ be the twelve real numbers that satisfy the system of equations \eqref{eq:répCsbmDébut}-\eqref{eq:répCsbmFin}. We introduce the two potentials
\begin{align}
\psi_w(w) &= -r\gamma(w) - \frac{1}{2}\hat V_ww^2 + \left(\varsigma\sqrt{\hat Q_w}+u\hat m_w\right) w \label{eq:potentielW} \\
\psi_\mathrm{out}(h,\sigma;\bar t) &= -\bar tl(yh)-\frac{1}{2}\hat V_\sigma\sigma^2+ \left(\xi\sqrt{\hat Q_\sigma}+y\hat m_\sigma\right)\sigma \label{eq:potentielOut} \\
&\quad{}+\log\mathcal N\left(h|c\sigma+\lambda ym_\sigma+\sqrt{Q_\sigma}\zeta, V_\sigma\right)+\log\mathcal N\left(\sigma|\sqrt\mu ym_w+\sqrt{Q_w}\chi, V_w\right) \nonumber
\end{align}
where $\mathcal N(\cdot|m,V)$ is a scalar Gaussian density of mean $m$ and variance $V$. The parameter $\bar t\in\{0,1\}$ controls if a given node is revealed $\bar t=1$ or not $\bar t=0$. We introduce the extremizers of these potentials:
\begin{align}
w^* = \argmax_w\psi_w(w) \label{eq:maxPotW} \\
(h^*,\sigma^*) = \argmax_{h,\sigma}\psi_\mathrm{out}(h,\sigma;\bar t=1) \qquad (h^{'*},\sigma^{'*}) &= \argmax_{h,\sigma}\psi_\mathrm{out}(h,\sigma;\bar t=0)\ . \label{eq:maxPotOut}
\end{align}
Then the expected errors and accuracies of the GCN on the CSBM are
\begin{align}
& E_\mathrm{train} = \mathbb E_{y,\xi,\zeta,\chi} l(yh^*) && \mathrm{Acc}_\mathrm{train} = \mathbb E_{y,\xi,\zeta,\chi} \delta_{y=\sign(h^*)} \\
& E_\mathrm{test} = \mathbb E_{y,\xi,\zeta,\chi} l(yh^{'*}) && \mathrm{Acc}_\mathrm{test} = \mathbb E_{y,\xi,\zeta,\chi} \delta_{y=\sign(h^{'*})}\ . \label{eq:formuleAccErr}
\end{align}
$\Theta$ and $\hat\Theta$ satisfy the following system of equations:
{\small
\begin{align}
& m_w = \frac{1}{\alpha}\mathbb E_{u,\varsigma}\,uw^* && m_\sigma=\mathbb E_{y,\xi,\zeta,\chi}\,y\mathcal P(\sigma) \label{eq:répCsbmDébut} \\
& Q_w=\frac{1}{\alpha}\mathbb E_{u,\varsigma}(w^*)^2 && Q_\sigma=\mathbb E_{y,\xi,\zeta,\chi}\mathcal P(\sigma^2) \\
& V_w=\frac{1}{\alpha}\frac{1}{\sqrt{\hat Q_w}}\mathbb E_{u,\varsigma}\,\varsigma w^* && V_\sigma=\frac{1}{\sqrt{\hat Q_\sigma}}\mathbb E_{y,\xi,\zeta,\chi}\,\xi\mathcal P(\sigma) \\
& \hat m_w=\frac{\sqrt\mu}{V_w}\mathbb E_{y,\xi,\zeta,\chi}\,y\mathcal P(\sigma-\sqrt\mu ym_w) && \hat m_\sigma=\frac{\lambda}{V_\sigma}\mathbb E_{y,\xi,\zeta,\chi}\,y\mathcal P(h-c\sigma-\lambda ym_\sigma) \\
& \hat Q_w=\frac{1}{V_w^2}\mathbb E_{y,\xi,\zeta,\chi}\mathcal P\left((\sigma-\sqrt\mu ym_w-\sqrt Q_w\chi)^2\right) && \hat Q_\sigma=\frac{1}{V_\sigma^2}\mathbb E_{y,\xi,\zeta,\chi}\mathcal P\left((h-c\sigma-\lambda ym_\sigma-\sqrt Q_\sigma\zeta)^2\right) \\
& \hat V_w=\frac{1}{V_w}\left(1-\frac{1}{\sqrt{Q_w}}\mathbb E_{y,\xi,\zeta,\chi}\,\chi\mathcal P(\sigma)\right) && \hat V_\sigma=\frac{1}{V_\sigma}\left(1-\frac{1}{\sqrt{Q_\sigma}}\mathbb E_{y,\xi,\zeta,\chi}\,\zeta\mathcal P(h-c\sigma)\right)\ . \label{eq:répCsbmFin}
\end{align}
}
For compactness we introduced the operator $\mathcal P$ that, for a polynomial $Q$ in $h$ and $\sigma$, acts according to
\begin{equation}
\mathcal P(Q(h,\sigma))=\rho Q(h^*,\sigma^*)+(1-\rho)Q(h^{'*},\sigma^{'*})\ .
\end{equation}
For instance $\mathcal P(\sigma^2)=\rho(\sigma^*)^2+(1-\rho)(\sigma^{'*})^2$.
\end{result}
The analysis of the GCN is thus reduced to the analysis of a finite set of scalar quantities $\Theta$ and $\hat\Theta$. They are called the summary statistics (or order parameters) of this model and they entirely describe its macroscopic properties. The equations \eqref{eq:répCsbmDébut}-\eqref{eq:répCsbmFin} they satisfy are called the self-consistent or fixed-point equations.

\begin{result}[Performances on the GLM--SBM]
The performances of the GCN on the GLM--SBM are given by the same formulae as for the CSBM, except that $\psi_\mathrm{out}$ is taken at $\mu=0$, that the law of $y$ is
\begin{equation}
P(y=\pm 1|\chi)=\frac{1}{2}\left(1\pm\erf\left(\frac{m_w\chi}{\sqrt{2(\alpha^{-1}Q_w-m_w^2)}}\right)\right)\ ,
\end{equation}
and that $\Theta$ and $\hat\Theta$ are the solution to the equations \eqref{eq:répGlmSbmDébut}-\eqref{eq:répGlmSbmFin}:
{\small
\begin{align}
& m_w = \frac{1}{\alpha}\mathbb E_{u,\varsigma}\,uw^* && m_\sigma=\mathbb E_{\xi,\zeta,\chi}\mathbb E_y\,y\mathcal P(\sigma) \label{eq:répGlmSbmDébut} \\
& Q_w=\frac{1}{\alpha}\mathbb E_{u,\varsigma}(w^*)^2 && Q_\sigma=\mathbb E_{\xi,\zeta,\chi}\mathbb E_y\mathcal P(\sigma^2) \\
& V_w=\frac{1}{\alpha}\frac{1}{\sqrt{\hat Q_w}}\mathbb E_{u,\varsigma}\,\varsigma w^* && V_\sigma=\frac{1}{\sqrt{\hat Q_\sigma}}\mathbb E_{\xi,\zeta,\chi}\mathbb E_y\,\xi\mathcal P(\sigma) \\
& \hat m_w=\frac{1}{V_w}\mathbb E_{\xi,\zeta,\chi}\sum_{y=\pm 1}yg(\chi)\mathcal P(\sigma) && \hat m_\sigma=\frac{\lambda}{V_\sigma}\mathbb E_{\xi,\zeta,\chi}\mathbb E_y\,y\mathcal P(h-c\sigma-\lambda ym_\sigma) \\
& \hat Q_w=\frac{1}{V_w^2}\mathbb E_{\xi,\zeta,\chi}\,\mathbb E_y\mathcal P\left((\sigma-\sqrt Q_w\chi)^2\right) && \hat Q_\sigma=\frac{1}{V_\sigma^2}\mathbb E_{\xi,\zeta,\chi}\mathbb E_y\mathcal P\left((h-c\sigma-\lambda ym_\sigma-\sqrt Q_\sigma\zeta)^2\right) \\
& \hat V_w=\frac{1}{V_w}\mkern-5mu\left(\mkern-3mu 1-\frac{1}{\sqrt{Q_w}}\mathbb E_{\xi,\zeta,\chi}\mkern-3mu\left(\mkern-3mu\mathbb E_y\,\chi\mathcal P(\sigma)\mkern-3mu - \mkern-5mu\sum_{y=\pm 1}\frac{ym_w}{\sqrt Q_w}g(\chi)\mathcal P(\sigma)\mkern-3mu\right)\mkern-4mu\right) && \hat V_\sigma=\frac{1}{V_\sigma}\left(1-\frac{1}{\sqrt{Q_\sigma}}\mathbb E_{\xi,\zeta,\chi}\mathbb E_y\,\zeta\mathcal P(h-c\sigma)\right)\ . \label{eq:répGlmSbmFin}
\end{align}
}
For compactness we introduced
\begin{equation}
g(\chi)=\frac{e^{-\frac{\eta_w}{2(1-\eta_w)}\chi^2}}{\sqrt{2\pi\alpha^{-1}(1-\eta_w)}} \quad\mathrm{and}\quad \eta_w=\alpha\frac{m_w^2}{Q_w}\ .
\end{equation}
\end{result}
In general there is no simple expression to the solution of the self-consistent equations and one has to solve them numerically or to consider special cases. We consider the limiting case $r\to\infty$. It is particularly relevant for two reasons. First in this limit simple explicit expressions can be stated; we give them in appendix~\ref{sec:pointFixe-rGrand} and in result~\ref{res:précisionsRinf}. Second, as we will show in \ref{sec:perteReg}, it corresponds to the optimal performance of the GCN on the CSBM, and close to optimal for the GLM--SBM, and it is thus the right limit to analyze how effective the GCN is. The ridge-less limit $r=0$ and $\alpha\rho>1$ has been studied by \cite{shi2022statistical} for the CSBM. We checked that in this case our expressions for the errors and the accuracies match theirs.

\begin{result}[Large regularization case]
\label{res:précisionsRinf}
We consider $r\to\infty$. For simplicity we state here the case $c=0$; the case $c\neq 0$ is given in appendix~\ref{sec:pointFixe-rGrand}. Then the test accuracy of the trained GCN is
\begin{align}
\mathrm{Acc}_\mathrm{test} &= \frac{1}{2}\left(1+\erf(\lambda\sqrt{\tau})\right)
\end{align}
where $\tau$ reads, respectively on the CSBM and on the GLM--SBM:
\begin{align}
\sqrt{\tau_\mathrm{CSBM}} &= \frac{\lambda\rho(1+\mu)}{\sqrt 2\sqrt{\rho(1+\alpha)+\lambda^2\rho^2(1+\mu)(1+\alpha+\mu)}} \label{eq:tauxCsbm} \\
\sqrt{\tau_\mathrm{GLM-SBM}} &= \frac{\lambda\rho(1+2\alpha/\pi)}{\sqrt 2\sqrt{\rho(1+\alpha)+\lambda^2\rho^2((1+2\alpha/\pi)(1+\alpha+2\alpha/\pi)-4\alpha^2/\pi^2)}} \label{eq:tauxGlmSbm}
\end{align}
\end{result}

\paragraph{Outline of the derivation:}
We compute the expected errors and accuracies in the high-dimensional limit $N$ and $M$ large. This problem can be phrased in the same way as in \cite{shi2022statistical}. We define an extended loss function (the Hamiltonian)
\begin{equation}
H(w) = t\sum_{i\in R}l(y_ih(w)_i)+r\sum_\nu\gamma(w_\nu) + t'\sum_{i\in R'}l(y_ih(w)_i)
\end{equation}
where $t$ and $t'$ are external parameters to probe the observables. The loss of the test samples is in $H$ for the purpose of the analysis; we will take $t'=0$ later and the algorithm is still minimizing the training loss eq.~\ref{eq:loss}. The moment generating function $f$ (the free energy) is defined as
\begin{equation}
Z = \int\mathrm dw\, e^{-\beta H(w)}\ ,\quad f = -\frac{1}{\beta N}\mathbb E\log Z\ .
\end{equation}
$\beta$ is an ancillary parameter (the inverse temperature) to minimize the loss: we consider the limit $\beta\to\infty$ where $Z$ (the partition function) concentrates over $w^*$ at $t=1$ and $t'=0$. The train and test errors are then obtained according to
\begin{equation}
E_\mathrm{train} = \frac{1}{\rho}\partial_t f \qquad\mathrm{and}\quad E_\mathrm{test} = \frac{1}{\rho'}\partial_{t'} f
\end{equation}
both evaluated at $t=1$ and $t'=0$. One can in the same manner compute the average accuracies by introducing the observables $\sum_{i\in\hat R}\delta_{y_i=\sign{h(w)_i}}$ in $H$.

To compute $f$ we use the powerful but non-rigorous replica method from Statistical Physics:
\begin{equation}
\mathbb E\log Z = \mathbb E\frac{\partial Z^n}{\partial n}(n=0) = \left(\frac{\partial}{\partial n}\mathbb EZ^n\right)(n=0)\ . \label{eq:replica}
\end{equation}
$Z^n$ is interpreted as having $n$ independent replica of the initial system, that become coupled by the expectation. We pursue the computation under the replica symmetry (RS) assumption, which is justified by the convexity of $H$. We introduce an intermediate variable $\sigma=\frac{1}{\sqrt N}Xw$ that corresponds to the projected features and that appears in the previous equations. The computation is then detailed in appendix \ref{sec:répliques}.

\begin{figure*}[t]
 \centering
 \includegraphics[width=0.95\linewidth]{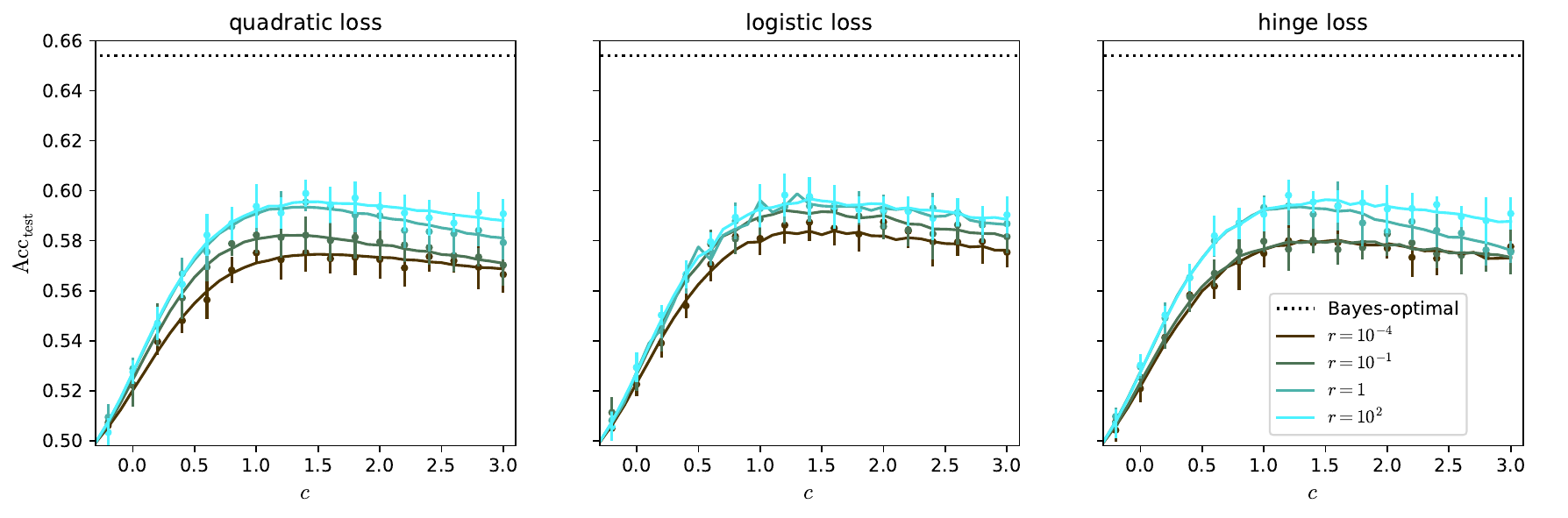}
 \includegraphics[width=0.95\linewidth]{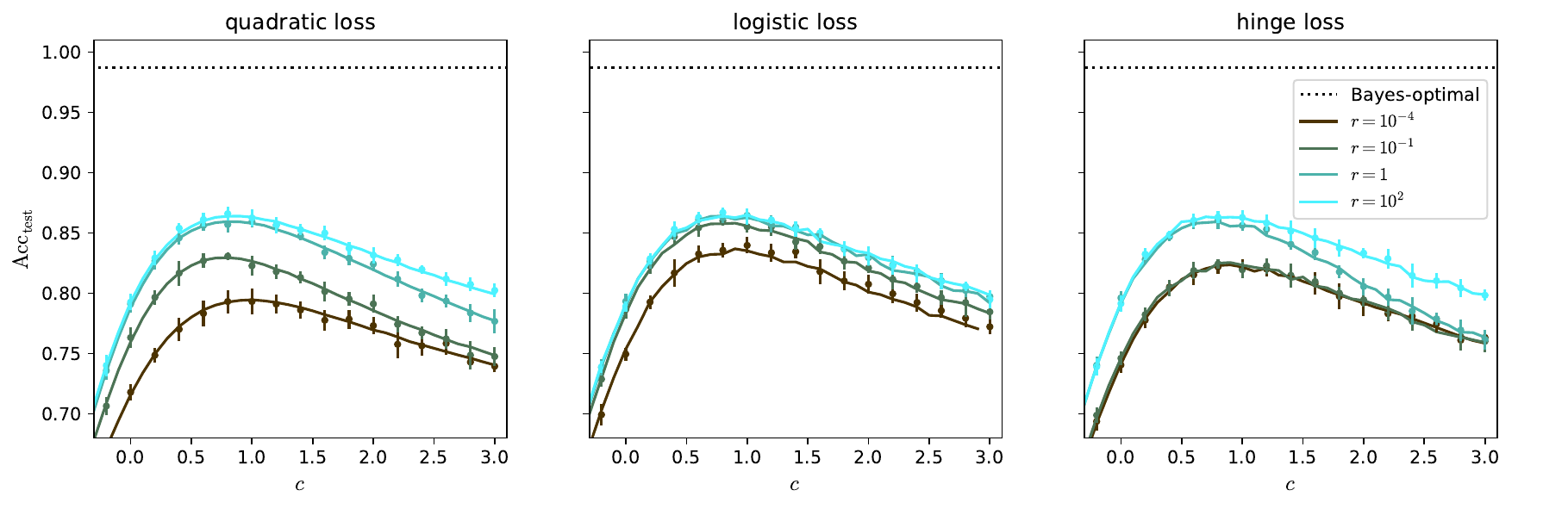}
 \caption{\label{fig:vsCR-csbm} Search for the optimal parameters of the GCN on CSBM. $\alpha=4$, $\rho=0.1$. \emph{Top:} low snr, $\lambda=0.5$, $\mu=1$. \emph{Bottom:} high snr, $\lambda=1.5$, $\mu=3$. Full lines: prediction for the test accuracy obtained by eqs.~\eqref{eq:formuleAccErr} and \eqref{eq:répCsbmDébut}-\eqref{eq:répCsbmFin}; dots: numerical simulation of the GCN for $N=10^4$ and $d=30$, averaged over ten experiments; dotted line: Bayes-optimal test accuracy.}
\end{figure*}

\section{Consequences}
\label{sec:consequences}
In the previous part we described the performances of the trained GCN by a finite set of summary statistics in the high-dimensional limit and we gave some explicit expressions. In this section we derive consequences from these equations. In particular we will search for the parameters of the GCN that optimize the test accuracy, to see whether the GCN can reach the Bayes-optimality. The possible tunable parameters are the self-loop intensity $c$, the regularization strength $r$ and the loss $l$. As to the regularization $\gamma$, we consider only $l_2$-regularization since we are in a simple setting not involving sparsity or outliers where $l_1$-regularization would have been beneficial. In general the system of equations \eqref{eq:répCsbmDébut}-\eqref{eq:répCsbmFin} and \eqref{eq:répGlmSbmDébut}-\eqref{eq:répGlmSbmFin} defining $\Theta$ and $\hat\Theta$ has to be solved numerically and one has to choose particular values for the parameters of the data models. For these, we consider both low and high snr, on both the CSBM and the GLM--SBM; we keep the signals of the graph and the features balanced and we take $\rho=0.1$ to mimic the common case where relatively few train labels are available. We did not explore all the parameters of the data models; instead we focused on plausible values and some corner cases may not follow our statements.

\begin{figure*}[t]
 \centering
 \includegraphics[width=0.9\linewidth]{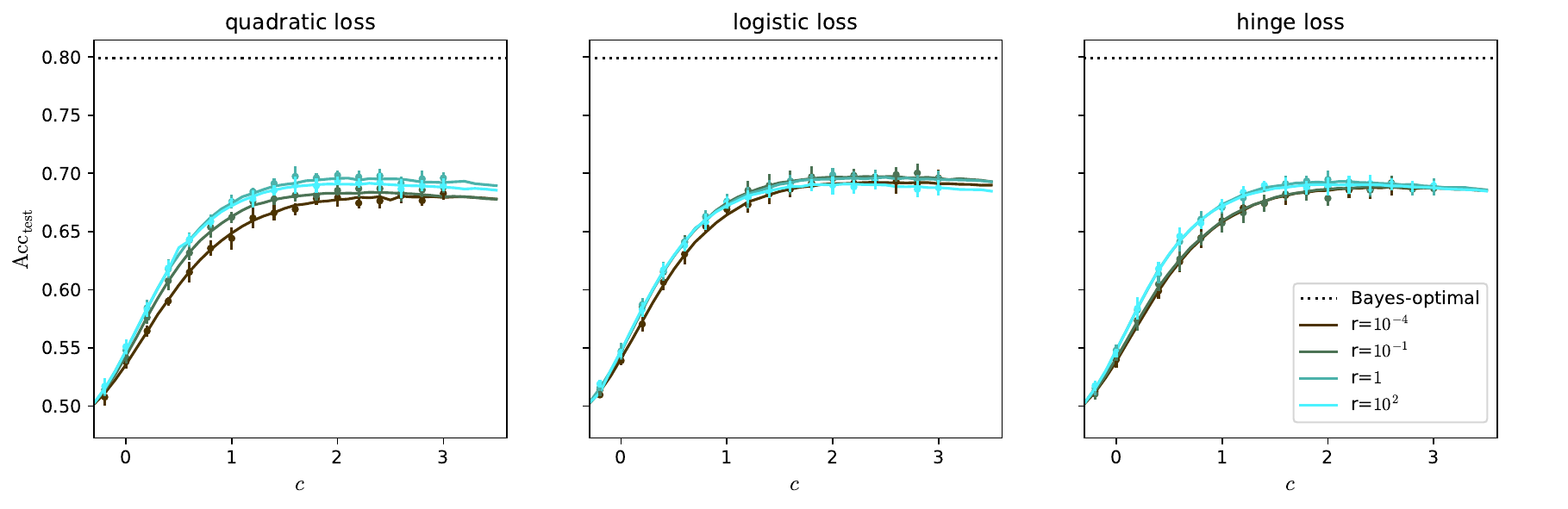}
 \includegraphics[width=0.9\linewidth]{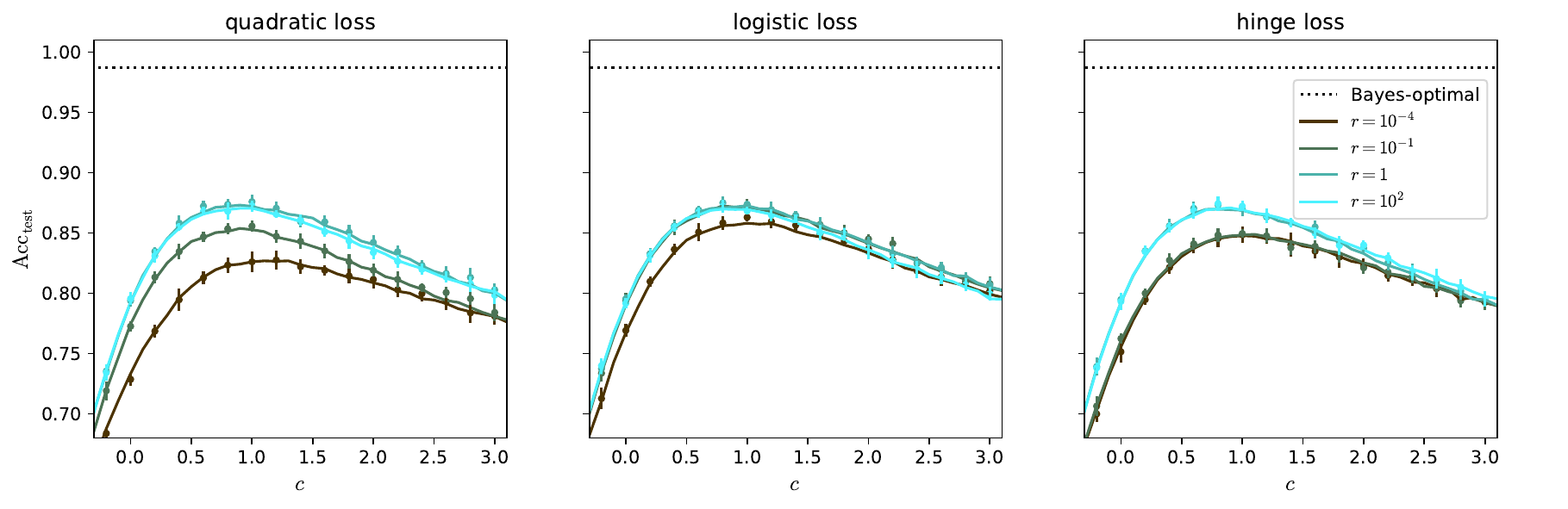}
 \caption{\label{fig:vsCR-glmSbm} Search for the optimal parameters of the GCN on GLM--SBM. $\alpha=4$, $\rho=0.1$. \emph{Top:} low snr, $\lambda=0.5$. \emph{Bottom:} high snr, $\lambda=1.5$. Full lines: prediction for the test accuracy obtained by eqs.~\eqref{eq:formuleAccErr} and \eqref{eq:répGlmSbmDébut}-\eqref{eq:répGlmSbmFin}; dots: numerical simulation of the GCN for $N=10^4$ and $d=30$, averaged over ten experiments; dotted line: Bayes-optimal test accuracy.}
\end{figure*}

Details on the numerics are provided in appendix~\ref{sec:numériques}. Our theoretical predictions are compared to simulations of the GCN on figs.~\ref{fig:vsCR-csbm}, \ref{fig:vsCR-glmSbm}, \ref{fig:vsCR-alpha0.7}, \ref{fig:vsCR-alpha2}, \ref{fig:picsBis} and \ref{fig:pics} for $N=10^4$ and $d=30$ or $d=N/2$. As expected, the predicted test accuracy, train accuracy and errors are within the statistical errors.

\begin{result}[Effect of the loss and the regularization]
\label{sec:perteReg}
Based on the numerical exploration of our equations shown in figs.~\ref{fig:vsCR-csbm}, \ref{fig:vsCR-glmSbm} and in figs.~\ref{fig:vsCR-alpha0.7} and \ref{fig:vsCR-alpha2} in appendix~\ref{sec:figuresSup}, we reach the conclusion that for both the CSBM and the GLM--SBM:
\begin{enumerate}
\item the optimal test accuracies $\mathrm{Acc}_\mathrm{test}$ depend little on the choice of the loss $l$. On the CSBM it appears to be reached at large regularization $r\to\infty$; on the GLM--SBM large regularization $r\to\infty$ is close to the optimal $r$;
\item there is an optimal self-loop strength $c^*$ maximizing $\mathrm{Acc}_\mathrm{test}$; $c^*$ is of order one;
\item there is a gap between the optimal test accuracy of the GCN and the Bayes-optimal test accuracy.
\end{enumerate}
\end{result}
We observe in figs.~\ref{fig:vsCR-csbm}, \ref{fig:vsCR-alpha0.7} and \ref{fig:vsCR-alpha2} that on the CSBM for all self-loop strengths $c$ the test accuracy increases with the regularization $r$ and reaches an optimal value at $r\to\infty$. As to the GLM--SBM, we observe in figs.~\ref{fig:vsCR-glmSbm}, \ref{fig:vsCR-alpha0.7} and \ref{fig:vsCR-alpha2} that $r\to\infty$ is close to the optimality, in particluar if $c$ is not too large.
Notice that at $r\to\infty$ the weights $w$ and the output $h(w)$ shrink to zero and that the test and train errors are large; yet this is not an issue: to assess the performance in a classification problem, the relevant quantity is the accuracy, not the error. At $r\to\infty$ the signs of $h(w)$ are mostly correct and the accuracies have a non-trivial value.
For both models the optimal $c$ is close to 1; this is consistent with \cite{shi2022statistical} that shows $c$ positive improves inference on homophilic graphs $\lambda>0$.
At low regularization $r$ we checked that interpolation peaks appear for the different losses while varying $\alpha$ or $\rho$; see figs.~\ref{fig:picsBis} and \ref{fig:pics} in appendix~\ref{sec:figuresSup}. Increasing $r$ smooths the peaks out, as \cite{shi2022statistical} shows for the quadratic loss; and as it is well known for the feed-forward networks, see e.g. \cite{mignacco20gaussMixt}.

A surprising result is that the optimal accuracy does not depend significantly on the loss; in particular, we do not see any significant difference between the three considered losses at optimal regularization. This is striking because it is rather generically anticipated that for classification the quadratic loss is less suitable than the logistic or hinge losses.
Indeed, in the feed-forward setting, \cite{aubin20glmReg} showed that the optimally regularized logistic regression improves significantly on the ridge regression. We do not observe such improvement in the present single-layer GCN setting where the features $X$ are mixed by the convolution $Q(\tilde A)X$. One previous example of $r\to\infty$ being optimal is classification on a binary high-dimensional Gaussian mixture \cite{mignacco20gaussMixt}. On the CSBM the CGN behaves similarly, which could be expected since the features $X$ are a Gaussian mixture. On the GLM--SBM where $X$ is generated by a GLM, it seems that they are partly mixed by the convolution $Q(\tilde A)X$, depending on the self-loops $c$. The fact that at $r\to\infty$ the three losses behave similarly is expected because the output $h$ is small and $l$ can be expanded around 0, where the three losses are identical.

More generally, at fixed small $r$, the logistic/hinge loss has better performances than the quadratic loss, as shown on figs.~\ref{fig:vsCR-csbm} and \ref{fig:vsCR-glmSbm}. If not regularized the quadratic loss always suffers from the interpolation peak at $\rho\alpha=1$, where the test accuracy is $1/2$, as shown on fig.~\ref{fig:picsBis}. For the logistic/hinge loss, the interpolation threshold is less harmful and it can be moved away with $\lambda$ and $c$, as shown on fig.~\ref{fig:pics}. A consequence is that at large $\lambda$ the logistic/hinge loss does not need regularization and reaches its optimal value even at small $r$, as depicted on fig.~\ref{fig:vsLambdaGrandVsR} in app.~\ref{sec:figuresSup}, while the quadratic loss needs $r\to\infty$.

Another remarkable point is that the performances of the GCN are far from the Bayes-optimal performances (dotted lines in the figures) in all cases. This is a major difference with the feed-forward case \cite{aubin20glmReg,mignacco20gaussMixt}, which shows that well-regularized regression performs very closely to the Bayes-optimal accuracy.
One could argue that this can be expected since the GCN performs only one step of convolution; estimators $Q(\tilde A)Xw$ with a higher-order polynomial $Q$ could be better. Yet such a gap exists even for more elaborated GNNs on CSBM \cite{dz23csbm} and GLM--SBM \cite{dz23glmSbm}.

\begin{figure}[t]
 \centering
 \includegraphics[width=0.47\linewidth]{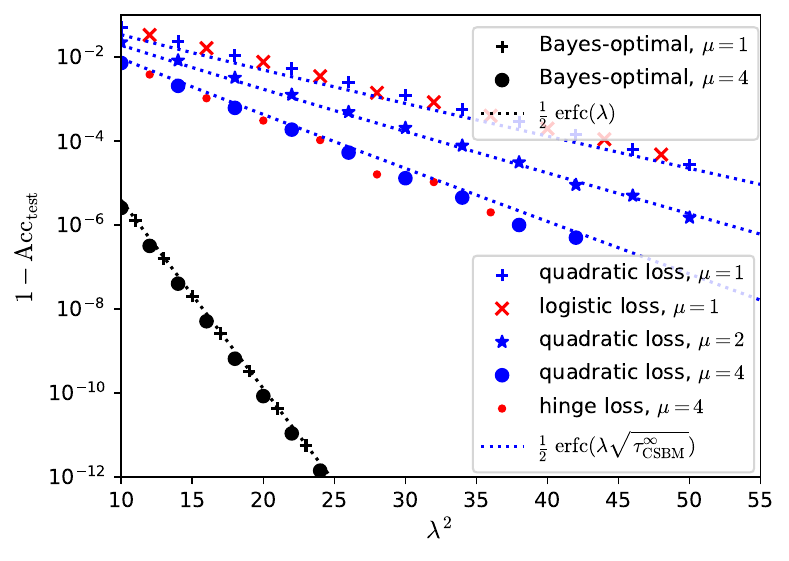}
 \includegraphics[width=0.47\linewidth]{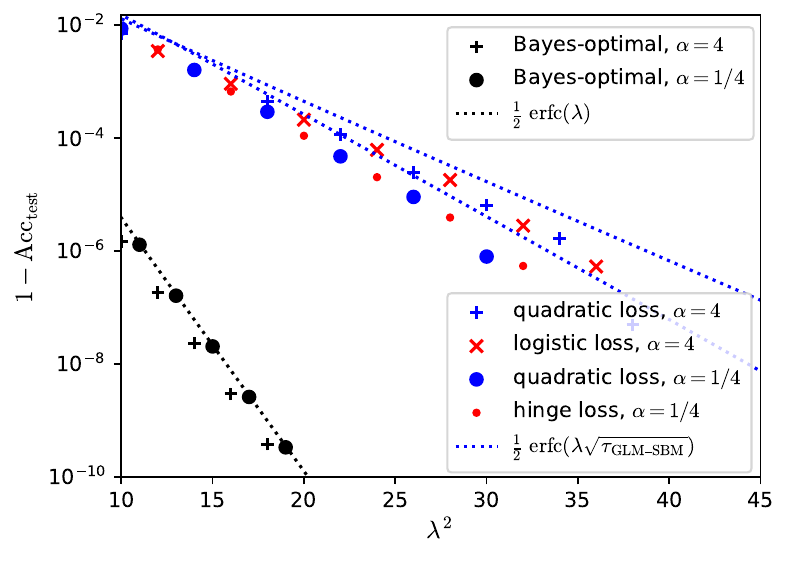}
 \caption{\label{fig:vsLambdaGrand} Asymptotic misclassification error $1-\mathrm{Acc}_\mathrm{test}$; \emph{left:} on the CSBM, $\alpha=4$; \emph{right:} on the GLM--SBM. $r=10^3$, $\rho=0.1$. Dots: prediction for the test accuracy obtained by eqs.~\eqref{eq:formuleAccErr}, \eqref{eq:répCsbmDébut}-\eqref{eq:répCsbmFin} and \eqref{eq:répGlmSbmDébut}-\eqref{eq:répGlmSbmFin}, for $c=c^*$ optimal obtained by grid search. Dotted lines are given by \eqref{eq:tauxLgrand} (for $\tau_\mathrm{CSBM}^\infty$) and \eqref{eq:tauxGlmSbm} (for $\tau_\mathrm{GLM-SBM}$). The Bayes-optimal values are obtained from the equations given in appendix~\ref{sec:eqBO}.}
\end{figure}

The two following results \ref{sec:consistanceTaux} and \ref{sec:cOptimal} come from the analysis of eqs.~\eqref{eq:prec-csbm} and \eqref{eq:prec-glmSbm} in appendix~\ref{sec:pointFixe-rGrand} as to the CGN, and from eqs.~\eqref{eq:precBOcsbm} and \eqref{eq:precBOglmSbm} in appendix~\ref{sec:eqBO} as to the Bayes-optimal performances.

\begin{result}[Consistency and convergence rates]
\label{sec:consistanceTaux}
We consider the limit of high graph signal $\lambda\to\infty$ at large regularization $r\to\infty$. We take $c=0$ or $c=c^*$ the optimal self-loop strength. Then the GCN is consistent on both models:
\begin{align}
\mathrm{Acc}_\mathrm{test} \underset{\lambda\to\infty}{\longrightarrow} 1 \ , \quad\quad \log(1-\mathrm{Acc}_\mathrm{test}) \underset{\lambda\to\infty}{\sim} -\lambda^2\tau^\infty\ ,
\end{align}
where $\tau^\infty$ is the asymptotic convergence (or learning) rate; for the CSBM and the GLM--SBM respectively it reads
\begin{align}
\tau_\mathrm{CSBM}^\infty = \frac{1+\mu}{2(1+\alpha+\mu)} \ , \quad\quad
\tau_\mathrm{GLM-SBM}^\infty = \frac{1+2\alpha/\pi}{2\left(1+\alpha+\frac{2\alpha/\pi}{1+2\alpha/\pi}\right)}\ . \label{eq:tauxLgrand}
\end{align}
Optimizing $\mathrm{Acc}_\mathrm{test}$ on $c$ only leads to a sub-leading improvement compared to taking $c=0$.
In both models the Bayes-optimal rate is
$\tau_\mathrm{BO}^\infty=1$.
\end{result}
Consequently, the GCN never reaches the Bayes-optimal rate. These statements are in agreement with the numerics depicted in fig.~\ref{fig:vsLambdaGrand}.

The expressions of the convergence rates \eqref{eq:tauxLgrand} are simple enough to be interpreted. As to $\tau_\mathrm{CSBM}^\infty$, this expression highlights the importance of the features: even at large graph snr $\lambda$ the GCN relies on the snr $\mu$ of the features. Indeed $\tau_\mathrm{CSBM}^\infty$ is increasing with $\mu$, from $1/2(1+\alpha)$ at $\mu=0$ to $1/2$ at large $\mu$. As suggested by the expression of the snr of the CSBM \eqref{eq:snrCsbm-GlmSbm}, increasing $\alpha$ lowers the performance, since $\tau_\mathrm{CSBM}^\infty$ goes to zero for large $\alpha$.
The respective snrs $\mu$ and $\alpha$ do not contribute to $\tau_\mathrm{CSBM}^\infty$ in the same manner as in \eqref{eq:snrCsbm-GlmSbm} where only the ratio $\mu^2/\alpha$ matters. This is a sign that the GCN does not handle the features optimally. The GCN also seems not to handle the graph optimally. Indeed, the Bayes-optimal rate $\tau_\mathrm{BO}=1$ does not depend on the feature snr $\mu$: hence, the graph alone is sufficient to reach the Bayes-optimal rate. We see a strong similitude between $\tau_\mathrm{GLM-SBM}^\infty$ and $\tau_\mathrm{CSBM}^\infty$. It is as if the feature snr $\mu$ of the CSBM were equivalent to an effective feature snr $2\alpha/\pi$ for the GLM--SBM. This is consistent with the expressions of the feature snrs of the two models \eqref{eq:snrCsbm-GlmSbm}, that are $\mu^2/\alpha$ and $4\alpha/\pi^2$. As to $\tau_\mathrm{GLM-SBM}^\infty$, it converges to a finite value for large $\alpha$, contrary to $\tau_\mathrm{CSBM}^\infty$ that goes to zero. This could be expected since the snr of the GLM--SBM \eqref{eq:snrCsbm-GlmSbm} is increasing with $\alpha$. A less intuitive result is that $\tau_\mathrm{GLM-SBM}^\infty$ reaches its maximum for $\alpha$ going to zero, as for $\tau_\mathrm{CSBM}^\infty$. It seems that there is a trade-off between the feature snr from the GLM (increasing with $\alpha$) and the resulting feature snr of the convoluted features $Q(\tilde A)X$ (decreasing with $\alpha$).

We notice that none of these rates depend on the training ratio $\rho$. We can also use these expressions to predict the performance of the GCN on the canonical SBM without features. More precisely, the CSBM at $\mu=0$ corresponds to a SBM populated with random Gaussian features. The rate reached by the GCN is better when $\alpha=\frac{N}{M}\to 0$ i.e. when we take the dimension $M$ as large as possible.

The learning rates $\tau_\mathrm{CSBM}^\infty$ and $\tau_\mathrm{GLM-SBM}^\infty$ can be straightforwardly obtained by taking the limit in $\tau_\mathrm{CSBM}$ and $\tau_\mathrm{GLM-SBM}$. Though being computed for $c=0$ they correctly described the leading behaviour of the GCN at $c=c^*$ because optimizing on $c$ only leads to a sub-leading improvement in the limit $\lambda\to\infty$. This is shown in fig.~\ref{fig:vsLambdaGrand} where the predicted values follow the slopes given by the different rates up to a small constant shift. As anticipated, this figure also shows that the three different losses give equal performances and the same rates.

The behaviour of the learning rates with respect to $r$ is depicted on fig.~\ref{fig:vsLambdaGrandVsR} in appendix~\ref{sec:figuresSup}. For the logistic loss, $\tau^\infty$ does not visibly depend on $r$ and even for small $r$ it achieves its optimal performance; while for the quadratic loss $\tau^\infty$ increases with $r$ up to its limit $\tau_\mathrm{GLM-SBM}^\infty$ \eqref{eq:tauxLgrand}. As explained in section \ref{sec:perteReg}, this is because the interpolation peak is always present for the quadratic loss, while for the logistic loss at large $\lambda$ and $c=c^*$ it disappears.

In conclusion, fig.~\ref{fig:vsLambdaGrand} further illustrates that the GCN does not reach the Bayes-optimal rate. For all considered settings $\tau_\mathrm{CSBM}^\infty$ and $\tau_\mathrm{GLM-SBM}^\infty$ are bounded by $1/2$ while $\tau_\mathrm{BO}=1$. Moreover, the two $\tau^\infty$ reach their upper bound $1/2$ only for the feature snr $\mu$ diverging or $\alpha$ going to zero, which confirms that the considered GCN has a rather poor performance.

\begin{figure*}[t]
 \centering
 \includegraphics[width=0.95\linewidth]{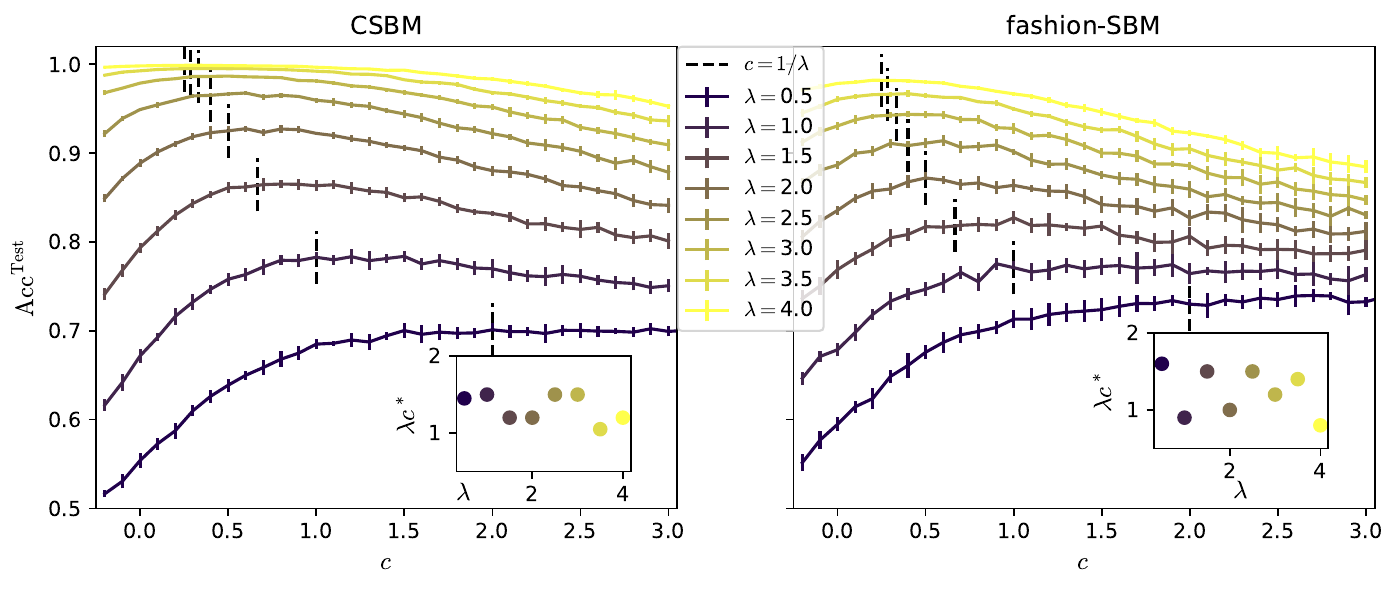}
 \caption{\label{fig:cOptimal} Optimal self-loop strength $c^*$ vs graph snr $\lambda$. $d=30$, $\rho=0.1$, $r=10^3$ and $l$ quadratic. \emph{Left:} on the CSBM, $N=10^4$, $\alpha=4$, $\mu=3$. \emph{Right:} on the fashion-SBM, classes 2 and 4. The lines are numerical simulations of the GCN averaged over ten experiments. $c^*$ is computed as the extremizer of the simulated $\mathrm{Acc}_\mathrm{test}$.}
\end{figure*}

\begin{result}[Optimal self-loop strength $c^*$]
\label{sec:cOptimal}
We consider the limit $r\to\infty$. At $\lambda\to 0$, the optimal self-loop strength $c^*$ reads
\begin{align}
c_\mathrm{CSBM}^* = \frac{\mu\left((1+\alpha)(2-\rho)+\rho(1+\mu)(1+\mu+\alpha)\right)}{\alpha(1+\mu)(2+\rho\mu)}\frac{1}{\lambda} \ , \quad\quad c_\mathrm{GLM-SBM}^* = \Theta(1/\lambda)\ . \label{eq:cOptimal-lPetit}
\end{align}
At $\lambda\to\infty$, the optimal self-loop strength $c^*$ reads
\begin{align}
c_\mathrm{CSBM}^* = \frac{1+\mu+\alpha}{\alpha}\frac{1}{\lambda} \ , \quad\quad c_\mathrm{GLM-SBM}^* = \Theta(1/\lambda) \label{eq:cOptimal-lGrand}
\end{align}
where for the GLM--SBM the constant is given by solving eq. \eqref{eq:cOpt-glmSbm}.
\end{result}
$c^*$ behaves like $1/\lambda$ for $\lambda$ both large and small and on both data models. Fig.~\ref{fig:cOptimal} left shows that $c^*$ can be approximated by $1/\lambda$ even for $\lambda$ of order one. Fig.~\ref{fig:cOptimal} right shows that the dependency $c^*\approx 1/\lambda$ seems to hold on a semi-realistic dataset, the fashion-SBM, defined in appendix~\ref{sec:fashionSBM}.

The case $\lambda\to\infty$ for the CSBM \eqref{eq:cOptimal-lGrand} is simple enough to be interpreted: $c^*_\mathrm{CSBM}$ increases with $\mu$ and decreases with $\alpha$; this means that the larger the feature snr is, the more the features should be taken in account in the convolution, which is expected. Conversely, $c^*$ increases when the graph snr $\lambda$ decreases and reaches $\infty$ when $\lambda=0$: the noisier the graph the less it should be considered in the convolution. The same happens in the case $\lambda\to 0$ for the CSBM \eqref{eq:cOptimal-lPetit} if $\rho$ is small, in which case we have $c^*_\mathrm{CSBM}=\mu(1+\alpha)/\lambda\alpha(1+\mu)$. For an arbitrary $\lambda$, for the two models, $c^*$ can still be predicted as the maximizer of eqs.~\eqref{eq:prec-csbm} or \eqref{eq:prec-glmSbm} in the appendix, but it does not admit a simple expression.

An interesting result is that $c^*$ behaves like $1/\lambda$ for $\lambda$ both small and large, for both models. Though the constant factors differ, this suggests a universal behaviour, for any $\lambda$ and beyond the two analyzed data models. We conjecture that, in general, taking $c^*=1/\lambda$ is a good approximation for the extremizer of the test accuracy. We tested this conjecture: first (fig.~\ref{fig:cOptimal} left) by considering $\lambda$ of order one on the CSBM, and second (fig.~\ref{fig:cOptimal} right) by training the GCN on a semi-realistic data model, the fashion-SBM, for which the features are taken from the fashion-MNIST dataset \cite{fashionMNIST17}. Fashion-SBM is defined defined in appendix \ref{sec:fashionSBM}. In the two cases, for $\lambda$ ranging from 0.5 to 4 we observe that $\lambda c^*$ remains close to 1, which seems to confirm our conjecture. This suggests that the rule $c^*=1/\lambda$ can be extended to a broader range of data, not only from the CSBM or the GLM--SBM, and could be useful in practice. A theoretical interpretation of this universality could be that the convolution $Q(\tilde A)X$ tends to transform the features $X$ to a Gaussian mixture, irrespectively to their distribution. This would explain why the same behaviour appears for the different datasets.

\section{Conclusion}
We theoretically predicted the generalization performances and the optimal architecture of a one-layer GCN on two models of attributed graphs. We showed that the optimal test accuracy is achieved for a finite value of the self-loop intensity at large regularization; it does not depend visibly on the training loss and there is a significant gap to the Bayes-optimality. This stands both when the features and the labels are generated by a Gaussian mixture and when they are generated by a GLM. We derived the optimal learning rates of the GCN and showed they can be interpreted in terms of feature signal-to-noise ratios. The GCN is consistent at large graph snr but does not reach the Bayes-optimal rate. We hope this simple setting will be usefull in understanding which aspects of the GCN are key to reach the optimality.

A future direction of work could be to analyze more complex GNNs such as a GCN with higher-order graph convolution $Q(\tilde A)$ or an attention-based GNN and to see if they can reach the optimality. Another direction could be given by the work \cite{nandy23regressionReseau} that proposes a model for genes where the components of the features are correlated according to a graph. One could study the role of graph-induced regularization.

\section*{Acknowledgements}
We acknowledge discussions with Cheng Shi.
This work is supported by the Swiss National Science Foundation under grant SNFS SMArtNet (grant number 212049).

\bibliography{main}
\bibliographystyle{plain}

\newpage
\appendix

\section{Replica computation}
\label{sec:répliques}
In this appendix, we derive the equations given in section \ref{sec:theory} of the main text. 

\subsection{Gaussian equivalence}
\label{sec:équGauss}
To average over the adjacency matrix $\tilde A$ we rely on a Gaussian equivalence property. It states that the rescaled adjacency matrix $\tilde A_{ij} = \left(\frac{d}{N}\left(1-\frac{d}{N}\right)\right)^{-1/2}\left(A_{ij}-\frac{d}{N}\right)$ can be approximated by the rank-one plus noise matrix $A^\mathrm{g} = \frac{\lambda}{\sqrt N}yy^T+\Xi$ without changing the expected losses and accuracies of the model, in the limit of large average degree $d$. It has been stated in \cite{lesieur2017constrained} for the SBM, proved in \cite{cSBM18} as to the mutual information and tested in \cite{shi2022statistical} for the GCN for $d=\Theta(\sqrt N)$. In practice taking $d\gtrsim 20$ at $N=1000$ is enough to observe no difference for the losses and accuracies and assuming $d=\omega(1)$ should be sufficient.

For the equivalence property to hold, the GCN has to compute the convolution over $\tilde A$. The constant shift by $d/N$ can be interpreted as centering $A$ while the constant scaling by $\left(\frac{d}{N}\left(1-\frac{d}{N}\right)\right)^{-1/2}$ normalizes its variance. The convolution over $\tilde A$ can still be interpreted as a graph convolution. The scaling can be absorbed in $w$ and $r$; and if the graph is not too dense $d=o(N)$ the shift is negligeable.

\subsection{CSBM}
We first derive the results for the CSBM, generalizing the results of \cite{shi2022statistical} to arbitrary convex loss and regularization. As stated in eq.~\ref{eq:replica}, we introduce $n$ replica:
\begin{align}
& Z = \int\prod_\nu^M\mathrm dw_\nu P_W(w_\nu)e^{-\beta t\sum_{i\in R}l(y_ih(w)_i)-\beta t'\sum_{i\in R'}l(y_ih(w)_i)} \\
& -\beta Nf = \mathbb E_{u, \Xi, W, y}\log Z = \mathbb E_{u, \Xi, W, y}\frac{\partial}{\partial n}Z^n(n=0) \\
& \quad = \frac{\partial}{\partial n}(n=0) \underbrace{\mathbb E_{u, \Xi, W, y}\int\prod_a^n\prod_\nu^M\mathrm dw_\nu^aP_W(w_\nu^a)e^{\sum_a^n-\beta t\sum_{i\in R}l(y_ih(w^a)_i)-\beta t'\sum_{i\in R'}l(y_ih(w^a)_i)}}_*
\end{align}
where $P_W(w)=\exp(-\beta r\gamma(w))$ is the prior on the weights induced by the regularization. We introduce several ancillary variables via $\delta$-Dirac functions to decouple the random variables. We set $h=\frac{1}{\sqrt N}(A^\mathrm{g}+c\sqrt NI_N)\sigma$ and $\sigma=\frac{1}{\sqrt N}Xw$. Then we take the expectation on the Gaussian noise:
\begin{align}
* \propto &\, \mathbb E_{u, \Xi, W, y}\int \prod_{a,\nu}\mathrm dw_\nu^aP_W(w_\nu^a)\prod_{a,i}\mathrm dh_i^a\mathrm dq_i^a e^{-\beta t\sum_{a,i\in R}l(y_ih_i^a)-\beta t'\sum_{a,i\in R'}l(y_ih_i^a)+\sum_{a,i}\mathrm iq_i^a\left(h_i^a-h(w^a)_i\right)} \\
 = &\, \mathbb E_{u, \Xi, W, y}\int \prod_{a,\nu}\mathrm dw_\nu^aP_W(w_\nu^a)\prod_{a,i}\mathrm dh_i^a\mathrm dq_i^a\mathrm d\sigma_i^a\mathrm d\hat q_i^a e^{-\beta t\sum_{a,i\in R}l(y_ih_i^a)-\beta t'\sum_{a,i\in R'}l(y_ih_i^a)} \nonumber \\
 & \quad\quad e^{\sum_{a,i}\mathrm iq_i^a\left(h_i^a-\frac{1}{\sqrt N}\sum_j\left(c\sqrt N\delta_{i,j}+\frac{\lambda}{\sqrt N}y_iy_j+\Xi_{ij}\right)\sigma_j^a\right)+\sum_{a,i}\mathrm i\hat q_i^a\left(\sigma_i^a-\frac{1}{\sqrt N}\sum_\nu\left(\sqrt\frac{\mu}{N}y_ju_\nu+W_{j\nu}\right)w_\nu^a\right)} \\
 = &\, \mathbb E_{u, y}\int \prod_{a,\nu}\mathrm dw_\nu^aP_W(w_\nu^a)\prod_{a,i}\mathrm dh_i^a\mathrm dq_i^a\mathrm d\sigma_i^a\mathrm d\hat q_i^a e^{-\beta t\sum_{a,i\in R}l(y_ih_i^a)-\beta t'\sum_{a,i\in R'}l(y_ih_i^a)+\mathrm i\sum_{a,i}(q_i^ah_i^a+\hat q_i^a\sigma_i^a)} \nonumber \\
 & \quad\quad e^{-\mathrm i\sum_{a,i}\left(cq_i^a\sigma_i^a+\frac{\lambda}{N}y_iq_i^a\sum_jy_j\sigma_j^a\right)-\frac{1}{2N}\sum_{i,j,a,b}q_i^aq_i^b\sigma_j^a\sigma_j^b-\mathrm i\sum_{a,i}\frac{\sqrt\mu}{N}y_i\hat q_i^a\sum_\nu u_\nu w_\nu^a-\frac{1}{2N}\sum_{i,\nu,a,b}\hat q_i^a\hat q_i^bw_\nu^aw_\nu^b}\ .
\end{align}
We integrate over the $q$s and $\hat q$s. For simplicity we pack the replica into vectors of size $n$.
\begin{align}
* = &\, \mathbb E_{u, y}\int \prod_{a,\nu}\mathrm dw_\nu^aP_W(w_\nu^a)\prod_{a,i}\mathrm dh_i^a\mathrm dq_i^a\mathrm d\sigma_i^a\mathrm d\hat q_i^a e^{-\beta t\sum_{a,i\in R}l(y_ih_i^a)-\beta t'\sum_{a,i\in R'}l(y_ih_i^a)+\mathrm i\sum_iq_i^T\left(h_i-c\sigma_i-\frac{\lambda}{N}y_i\sum_jy_j\sigma_j\right)} \nonumber \\
 & \quad\quad e^{-\frac{1}{2N}\sum_iq_i^T\left(\sum_j\sigma_j\sigma_j^T\right)q_i+\mathrm i\sum_i\hat q_i^T\left(\sigma_i-\frac{\sqrt\mu}{N}y_i\sum_\nu u_\nu w_\nu\right)-\frac{1}{2N}\sum_i\hat q_i^T\left(\sum_\nu w_\nu w_\nu^T\right)\hat q_i} \\
 = &\, \mathbb E_{u, y}\int \prod_{a,\nu}\mathrm dw_\nu^aP_W(w_\nu^a)\prod_{a,i}\mathrm dh_i^a\mathrm d\sigma_i^a e^{-\beta t\sum_{a,i\in R}l(y_ih_i^a)-\beta t'\sum_{a,i\in R'}l(y_ih_i^a)} \nonumber \\
 & \quad \prod_i\mathcal N\left(h_i\left| c\sigma_i+\frac{\lambda}{N}y_i\sum_jy_j\sigma_j, \frac{1}{N}\sum_j\sigma_j\sigma_j^T\right.\right) \mathcal N\left(\sigma_i\left| \frac{\sqrt\mu}{N}y_i\sum_\nu u_\nu w_\nu, \frac{1}{N}\sum_\nu w_\nu w_\nu^T\right.\right)\ ;
\end{align}
where $\mathcal N(x|\mu,\Sigma)=\mathrm{det}(2\pi\Sigma)^{-1/2}e^{-(x-\mu)^T\Sigma^{-1}(x-\mu)/2}$ is a Gaussian density.
The order parameters are
\begin{align}
& m_w^a = \frac{1}{N}\sum_\nu u_\nu w_\nu^a && m_\sigma^a=\frac{1}{N}\sum_i y_i\sigma_i^a \\
& Q_w^{ab} = \frac{1}{N}\sum_\nu w_\nu^aw_\nu^b && Q_\sigma^{ab} = \frac{1}{N}\sum_i\sigma_i^a\sigma_i^b
\end{align}
We introduce them via new $\delta$-Dirac functions. We can factorize the $i$ and $\nu$ indices.
\begin{align}
* \propto &\, \mathbb E_{u, y}\int \prod_{a,\nu}\mathrm dw_\nu^aP_W(w_\nu^a)\prod_{a,i}\mathrm dh_i^a\mathrm d\sigma_i^a\prod_{a\le b}\mathrm d\hat Q_w^{ab}\mathrm dQ_w^{ab}\mathrm d\hat Q_\sigma^{ab}\mathrm dQ_\sigma^{ab}\prod_a\mathrm d\hat m_w^a\mathrm dm_w^a\mathrm d\hat m_\sigma^a\mathrm dm_\sigma^a e^{-\beta t\sum_{a,i\in R}l(y_ih_i^a)} \nonumber \\
 & \quad\quad e^{-\beta t'\sum_{a,i\in R'}l(y_ih_i^a)}\prod_{a\le b}e^{\hat Q_w^{ab}\left(NQ_w^{ab}-\sum_\nu w_\nu^aw_\nu^b\right)+\hat Q_\sigma^{ab}\left(NQ_\sigma^{ab}-\sum_i\sigma_i^a\sigma_i^b\right)} \prod_ae^{\hat m_w^a(Nm_w^a-\sum_\nu u_\nu w_\nu^a)+\hat m_\sigma^a(Nm_\sigma^a-\sum_i y_i\sigma_i^a)} \nonumber \\
 & \quad\quad \prod_i N\left(h_i\left| c\sigma_i+\lambda y_im_\sigma, Q_\sigma\right.\right) \mathcal N\left(\sigma_i\left| \sqrt\mu y_im_w, Q_w\right.\right) \\
 = &\, \int \prod_{a\le b}\mathrm d\hat Q_w^{ab}\mathrm dQ_w^{ab}\mathrm d\hat Q_\sigma^{ab}\mathrm dQ_\sigma^{ab}\prod_a\mathrm d\hat m_w^a\mathrm dm_w^a\mathrm d\hat m_\sigma^a\mathrm dm_\sigma^a \prod_{a\le b}e^{N(\hat Q_w^{ab}Q_w^{ab}+\hat Q_\sigma^{ab}Q_\sigma^{ab})} \prod_ae^{N(\hat m_w^am_w^a+\hat m_\sigma^am_\sigma^a)} \nonumber \\
 & \quad\quad \left[\mathbb E_u\int \prod_a\mathrm dw^ae^{\psi_w^{(n)}(w)} \right]^{N/\alpha}
\left[\mathbb E_y\int \prod_a\mathrm dh^a\mathrm d\sigma^a e^{\psi_\mathrm{out}^{(n)}(h,\sigma;t)} \right]^{\rho N}
\left[\mathbb E_y\int \prod_a\mathrm dh^a\mathrm d\sigma^a e^{\psi_\mathrm{out}^{(n)}(h,\sigma;t')} \right]^{\rho'N} \nonumber \\
 & \quad\quad \left[\mathbb E_y\int \prod_a\mathrm dh^a\mathrm d\sigma^a e^{\psi_\mathrm{out}^{(n)}(h,\sigma;0)} \right]^{(1-\rho-\rho')N} \ ;
\end{align}
where we defined
\begin{align}
\psi_w^{(n)}(w) &= \sum_a\log P_W(w^a)-\sum_{a\le b}\hat Q_w^{ab}w^aw^b-\sum_a\hat m_w^auw^a \\
 \psi_\mathrm{out}^{(n)}(h,\sigma;\bar t) &= -\beta \bar t\sum_al(yh^a)-\sum_{a\le b}\hat Q_\sigma^{ab}\sigma^a\sigma^b-\sum_a\hat m_\sigma^ay\sigma^a-\frac{1}{2}(h-c\sigma-\lambda ym_\sigma)^TQ_\sigma^{-1}(h-c\sigma-\lambda ym_\sigma) \nonumber \\
& \quad\quad{}-\frac{1}{2}\log\mathrm{det}\,Q_\sigma-\frac{1}{2}(\sigma-\sqrt\mu ym_w)^TQ_w^{-1}(\sigma-\sqrt\mu ym_w)-\frac{1}{2}\log\mathrm{det}\,Q_w\ .
\end{align}
We use the replica-symmetric ansatz: we set $\hat Q^{aa}=\frac{1}{2}\hat R$, $\hat Q^{ab}=-\hat Q$, $Q^{aa}=R$, $Q^{ab}=Q$, $\hat m^a=-\hat m$ and $m^a=m$. Since we will take the derivative wrt $n$ and send $n$ to zero we discard all the terms that are not proportionnal to $n$. We compute first that
\begin{align}
Q^{-1} &= \frac{1}{R-Q}I_n-\frac{Q}{(R-Q)^2}J_{n,n} + o(n) \\
\log\mathrm{det}\,Q &= n\frac{Q}{R-Q}+n\log(R-Q) + o(n)\ ;
\end{align}
where $J_{n,n}$ is the matrix filled with ones. We define the variances $V=R-Q$ and $\hat V=\hat R+\hat Q$. We introduce scalar Gaussian random variables $\xi$ and $\chi$ to decouple the replica and factorize them. Then
\begin{align}
* \propto &\, \int \mathrm d\hat Q_w\mathrm d\hat V_w\mathrm dQ_w\mathrm dV_w\mathrm d\hat Q_\sigma\mathrm d\hat V_\sigma\mathrm dQ_\sigma\mathrm dV_\sigma\mathrm d\hat m_w\mathrm dm_w\mathrm d\hat m_\sigma\mathrm dm_\sigma e^{\frac{nN}{2}(\hat V_wV_w+\hat V_wQ_w-V_w\hat Q_w+\hat V_\sigma V_\sigma+\hat V_\sigma Q_\sigma-V_\sigma\hat Q_\sigma)} \nonumber \\
 & \quad\quad e^{-nN(\hat m_wm_w+\hat m_\sigma m_\sigma)}\left[\mathbb E_{u,\xi}\left(\int \mathrm dw\,e^{\psi_w(w)}\right)^n \right]^{N/\alpha}
\left[\mathbb E_{y,\xi,\zeta,\chi}\left(\int \mathrm dh\mathrm d\sigma e^{\psi_\mathrm{out}(h,\sigma;t)}\right)^n \right]^{\rho N} \nonumber \\
 & \quad\quad \left[\mathbb E_{y,\xi,\zeta,\chi}\left(\int \mathrm dh\mathrm d\sigma e^{\psi_\mathrm{out}(h,\sigma;t')}\right)^n \right]^{\rho'N} \left[\mathbb E_{y,\xi,\zeta,\chi}\left(\int \mathrm dh\mathrm d\sigma e^{\psi_\mathrm{out}(h,\sigma;0)}\right)^n \right]^{(1-\rho-\rho')N} \\
:= &\, \int\mathrm dm\,\mathrm dq\,\mathrm dv\,e^{N\phi^{(n)}(m, q, v)}\ ,
\end{align}
with
\begin{align}
\psi_w(w) &= \log P_W(w)-\frac{1}{2}\hat V_ww^2+\left(\xi\sqrt{\hat Q_w}+u\hat m_w\right)w \\
 \psi_\mathrm{out}(h,\sigma;\bar t) &= -\beta \bar tl(yh)-\frac{1}{2}\hat V_\sigma\sigma^2+\left(\xi\sqrt{\hat Q_\sigma}+y\hat m_\sigma\right)\sigma \nonumber \\
 & \quad\quad {}+\log\mathcal N\left(h|c\sigma+\lambda ym_\sigma+\sqrt{Q_\sigma}\zeta, V_\sigma\right)
+\log\mathcal N\left(\sigma|\sqrt\mu ym_w+\sqrt{Q_w}\chi, V_w\right)
\end{align}
and $m$, $q$ and $v$ standing for all the order parameters. $\xi$, $\zeta$ and $\chi$ are scalar standard Gaussians.
We take the limit $N\to\infty$ thanks to Laplace's method.
\begin{align}
-\beta f\propto& \frac{1}{N}\frac{\partial}{\partial n}(n=0) \int\mathrm dm\,\mathrm dq\,\mathrm dv\,e^{N\phi^{(n)}(m, q, v)} \\
 =& \extr_{m, q, v}\frac{\partial}{\partial n}(n=0)\phi^{(n)}(m, q, v) := \extr_{m, q, v}\phi(m, q, v)\ ;
\end{align}
the free entropy is
\begin{align}
\phi &= \frac{1}{2}\left(\hat V_wV_w+\hat V_wQ_w-V_w\hat Q_w+\hat V_\sigma V_\sigma+\hat V_\sigma Q_\sigma-V_\sigma\hat Q_\sigma\right)-\hat m_wm_w-\hat m_\sigma m_\sigma \nonumber \\
 &\quad\quad {}+\frac{1}{\alpha}\mathbb E_{u,\xi}\left(\log\int\mathrm dw\,e^{\psi_w(w)}\right)+\rho\mathbb E_{y,\xi,\zeta,\chi}\left(\log\int\mathrm dh\mathrm d\sigma e^{\psi_\mathrm{out}(h,\sigma;t)}\right) \nonumber \\
 &\quad\quad {}+\rho'\mathbb E_{y,\xi,\zeta,\chi}\left(\log\int\mathrm dh\mathrm d\sigma e^{\psi_\mathrm{out}(h,\sigma;t')}\right) +(1-\rho-\rho')\mathbb E_{y,\xi,\zeta,\chi}\left(\log\int\mathrm dh\mathrm d\sigma e^{\psi_\mathrm{out}(h,\sigma;0)}\right)\ . 
\end{align}
We take the extremum of the free entropy deriving it wrt the order parameters, evaluated at $t=1$ and $t'=0$. We obtain the following fixed-point conditions.
\begin{align}
& m_w = \frac{1}{\alpha}\mathbb E_{u,\xi}\,u\,\mathbb E_{P_w}w \quad && m_\sigma=\mathbb E_{y,\xi,\zeta,\chi}\,y\left(\rho\mathbb E_{P_\mathrm{out}}\sigma+(1-\rho)\mathbb E_{P_\mathrm{out}'}\sigma\right) \\
& Q_w+V_w=\frac{1}{\alpha}\mathbb E_{u,\xi}\mathbb E_{P_w}w^2 \quad && Q_\sigma+V_\sigma=\mathbb E_{y,\xi,\zeta,\chi}\left(\rho\mathbb E_{P_\mathrm{out}}\sigma^2+(1-\rho)\mathbb E_{P_\mathrm{out}'}\sigma^2\right) \\
& V_w=\frac{1}{\alpha}\frac{1}{\sqrt{\hat Q_w}}\mathbb E_{u,\xi}\,\xi\,\mathbb E_{P_w}w \quad && V_\sigma=\frac{1}{\sqrt{\hat Q_\sigma}}\mathbb E_{y,\xi,\zeta,\chi}\,\xi\left(\rho\mathbb E_{P_\mathrm{out}}\sigma+(1-\rho)\mathbb E_{P_\mathrm{out}'}\sigma\right)
\end{align}
\begin{align}
& \hat m_w=\frac{\sqrt\mu}{V_w}\mathbb E_{y,\xi,\zeta,\chi}\,y\left(\rho\mathbb E_{P_\mathrm{out}}(\sigma-\sqrt\mu ym_w)+(1-\rho)\mathbb E_{P_\mathrm{out}'}(\sigma-\sqrt\mu ym_w)\right) \\
& \hat Q_w-\hat V_w=\frac{1}{V_w^2}\mathbb E_{y,\xi,\zeta,\chi}\left(\rho\mathbb E_{P_\mathrm{out}}(\sigma-\sqrt\mu ym_w-\sqrt Q_w\chi)^2+(1-\rho)\mathbb E_{P_\mathrm{out}'}(\sigma-\sqrt\mu ym_w-\sqrt Q_w\chi)^2\right) - \frac{1}{V_w}\\
& \hat V_w=\frac{1}{V_w}\left(1-\frac{1}{\sqrt{Q_w}}\mathbb E_{y,\xi,\zeta,\chi}\,\chi\left(\rho\mathbb E_{P_\mathrm{out}}\sigma+(1-\rho)\mathbb E_{P_\mathrm{out}'}\sigma\right)\right) \\
& \hat m_\sigma=\frac{\lambda}{V_\sigma}\mathbb E_{y,\xi,\zeta,\chi}\,y\left(\rho\mathbb E_{P_\mathrm{out}}(h-c\sigma-\lambda ym_\sigma)+(1-\rho)\mathbb E_{P_\mathrm{out}'}(h-c\sigma-\lambda ym_\sigma)\right) \\
& \hat Q_\sigma-\hat V_\sigma=\frac{1}{V_\sigma^2}\mathbb E_{y,\xi,\zeta,\chi}\left(\rho\mathbb E_{P_\mathrm{out}}(h-c\sigma-\lambda ym_\sigma-\sqrt Q_\sigma\zeta)^2+(1-\rho)\mathbb E_{P_\mathrm{out}'}(h-c\sigma-\lambda ym_\sigma-\sqrt Q_\sigma\zeta)^2\right) - \frac{1}{V_\sigma} \\
& \hat V_\sigma=\frac{1}{V_\sigma}\left(1-\frac{1}{\sqrt{Q_\sigma}}\mathbb E_{y,\xi,\zeta,\chi}\,\zeta\left(\rho\mathbb E_{P_\mathrm{out}}(h-c\sigma)+(1-\rho)\mathbb E_{P_\mathrm{out}'}(h-c\sigma)\right)\right)
\end{align}
The measures are
\begin{align}
\mathrm dP_w = \frac{\mathrm dw\,e^{\psi_w(w)}}{\int\mathrm dw\,e^{\psi_w(w)}} \quad,\quad
\mathrm dP_\mathrm{out} = \frac{\mathrm dh\mathrm d\sigma\,e^{\psi_\mathrm{out}(h,\sigma;\bar t=1)}}{\int\mathrm dh\mathrm d\sigma\,e^{\psi_\mathrm{out}(h,\sigma,\bar t=1)}} \quad,\quad
\mathrm dP_\mathrm{out}' = \frac{\mathrm dh\mathrm d\sigma\,e^{\psi_\mathrm{out}(h,\sigma;\bar t=0)}}{\int\mathrm dh\mathrm d\sigma\,e^{\psi_\mathrm{out}(h,\sigma,\bar t=0)}}\ .
\end{align}
These measures can be computed thanks to Laplace's method in the limit $\beta\to\infty$. We have to rescale the order parameters not to obtain a degenerated solution. We recall that $\log P_W(w)\propto\beta$. We take $\hat V\to\beta\hat V$, $\hat Q\to\beta^2\hat Q$, $\hat m\to\beta\hat m$ and $V\to\beta^{-1}V$ for both $w$ and $\sigma$. We define
\begin{align}
& w^* = \argmax_w\psi_w(w) && \\
& (h^*,\sigma^*) = \argmax_{h,\sigma}\psi_\mathrm{out}(h,\sigma;\bar t=1) && (h^{'*},\sigma^{'*}) = \argmax_{h,\sigma}\psi_\mathrm{out}(h,\sigma;\bar t=0)\ ;
\end{align}
then, keeping the first order in $\beta$ in both lhs and rhs, the fixed-point equations are
\begin{align}
& m_w = \frac{1}{\alpha}\mathbb E_{u,\xi}\,uw^* \quad && m_\sigma=\mathbb E_{y,\xi,\zeta,\chi}\,y\left(\rho\sigma^*+(1-\rho)\sigma^{'*}\right) \\
& Q_w=\frac{1}{\alpha}\mathbb E_{u,\xi}(w^*)^2 \quad && Q_\sigma=\mathbb E_{y,\xi,\zeta,\chi}\left(\rho(\sigma^*)^2+(1-\rho)(\sigma^{'*})^2\right) \\
& V_w=\frac{1}{\alpha}\frac{1}{\sqrt{\hat Q_w}}\mathbb E_{u,\xi}\,\xi w^* \quad && V_\sigma=\frac{1}{\sqrt{\hat Q_\sigma}}\mathbb E_{y,\xi,\zeta,\chi}\,\xi\left(\rho\sigma^*+(1-\rho)\sigma^{'*}\right)
\end{align}
\begin{align}
& \hat m_w=\frac{\sqrt\mu}{V_w}\mathbb E_{y,\xi,\zeta,\chi}\,y\left(\rho(\sigma^*-\sqrt\mu ym_w)+(1-\rho)(\sigma^{'*}-\sqrt\mu ym_w)\right) \\
& \hat Q_w=\frac{1}{V_w^2}\mathbb E_{y,\xi,\zeta,\chi}\left(\rho(\sigma^*-\sqrt\mu ym_w-\sqrt Q_w\chi)^2+(1-\rho)(\sigma^{'*}-\sqrt\mu ym_w-\sqrt Q_w\chi)^2\right) \\
& \hat V_w=\frac{1}{V_w}\left(1-\frac{1}{\sqrt{Q_w}}\mathbb E_{y,\xi,\zeta,\chi}\,\chi\left(\rho\sigma^*+(1-\rho)\sigma^{'*}\right)\right) \\
& \hat m_\sigma=\frac{\lambda}{V_\sigma}\mathbb E_{y,\xi,\zeta,\chi}\,y\left(\rho(h^*-c\sigma^*-\lambda ym_\sigma)+(1-\rho)(h^{'*}-c\sigma^{'*}-\lambda ym_\sigma)\right) \\
& \hat Q_\sigma=\frac{1}{V_\sigma^2}\mathbb E_{y,\xi,\zeta,\chi}\left(\rho(h^*-c\sigma^*-\lambda ym_\sigma-\sqrt Q_\sigma\zeta)^2+(1-\rho)(h^{'*}-c\sigma^{'*}-\lambda ym_\sigma-\sqrt Q_\sigma\zeta)^2\right) \\
& \hat V_\sigma=\frac{1}{V_\sigma}\left(1-\frac{1}{\sqrt{Q_\sigma}}\mathbb E_{y,\xi,\zeta,\chi}\,\zeta\left(\rho(h^*-c\sigma^*)+(1-\rho)(h^{'*}-c\sigma^{'*})\right)\right)
\end{align}

The average train and test losses can be computed by deriving $\phi$ wrt $t$ and $t'$ and taking it extremum by evaluating it at the fixed-point of these equations. Simplifying the notations we obtain the equations given in the main part.

\subsection{GLM--SBM}
We derive the results for the GLM--SBM, which has not been studied by \cite{shi2022statistical}.
The derivation is similar to the derivation of the previous part on the CSBM. As we saw for the CSBM, one can readily take the test set $R'$ being the complement of $R$ i.e. $\rho'=1-\rho$; the resulting equations do not change. As stated in eq.~\ref{eq:replica}, we introduce $n$ replica:
\begin{align}
& Z = \mkern-5mu\int\mkern-3mu \prod_\nu^M\mathrm dw_\nu P_W(w_\nu)\prod_i^N\mathrm dy_iP_o\mkern-3mu\left(y_i|\frac{1}{\sqrt N}X_i^Tu\right)e^{-\beta t\sum_{i\in R}l(y_ih(w)_i)-\beta t'\sum_{i\in R'}l(y_ih(w)_i)} \\
& -\beta Nf = \mathbb E_{u, \Xi, X}\log Z=\mathbb E_{u, \Xi, X}\frac{\partial}{\partial n}Z^n(n=0) = \frac{\partial}{\partial n}(n=0)\\
& \quad \underbrace{\mathbb E_{u, \Xi, X}\int\prod_a^n\prod_\nu^M\mathrm dw_\nu^aP_W(w_\nu^a)\prod_i^N\mathrm dy_iP_o\left(y_i|\frac{1}{\sqrt N}X_i^Tu\right)e^{\sum_a^n-\beta t\sum_{i\in R}l(y_ih(w^a)_i)-\beta t'\sum_{i\in R'}l(y_ih(w^a)_i)}}_* \nonumber
\end{align}
where $P_o(y|z)=\delta_{y=\sign(z)}$. We introduce ancillary variables: $h=\frac{1}{\sqrt N}(A^\mathrm{g}+c\sqrt NI_N)\sigma$, $\sigma=\frac{1}{\sqrt N}Xw$ and $z=\frac{1}{\sqrt N}Xu$; we average over $\Xi$ and $X$, pack the replica and integrate.
\begin{align}
* \propto &\, \mathbb E_{u, \Xi, X}\int \prod_{a,\nu}\mathrm dw_\nu^aP_W(w_\nu^a)\prod_i\mathrm dy_iP_o(y_i|z_i)\mathrm dz_i\mathrm d\bar q_i\prod_{a,i}\mathrm dh_i^a\mathrm dq_i^a\mathrm d\sigma_i^a\mathrm d\hat q_i^a e^{-\beta t\sum_{a,i\in R}l(y_ih_i^a)-\beta t'\sum_{a,i\in R'}l(y_ih_i^a)} \\
 & \quad e^{\sum_i\mathrm i\bar q_i\left(z_i-\frac{1}{\sqrt N}\sum_\nu X_{i\nu}u_\nu\right)+\sum_{a,i}\mathrm iq_i^a\left(h_i^a-\frac{1}{\sqrt N}\sum_j\left(c\sqrt N\delta_{i,j}+\frac{\lambda}{\sqrt N}y_iy_j+\Xi_{ij}\right)\sigma_j^a\right)+\sum_{a,i}\mathrm i\hat q_i^a\left(\sigma_i^a-\frac{1}{\sqrt N}\sum_\nu X_{i\nu}w_\nu^a\right)} \nonumber \\
 =&\, \mathbb E_{u}\int \prod_{a,\nu}\mathrm dw_\nu^aP_W(w_\nu^a)\prod_i\mathrm dy_iP_o(y_i|z_i)\mathrm dz_i\prod_{a,i}\mathrm dh_i^a\mathrm d\sigma_i^a e^{-\beta t\sum_{a,i\in R}l(y_ih_i^a)-\beta t'\sum_{a,i\in R'}l(y_ih_i^a)} \\
 & \quad \prod_i\mathcal N\left(h_i\left| c\sigma_i+\frac{\lambda}{N}y_i\sum_jy_j\sigma_j, \frac{1}{N}\sum_j\sigma_j\sigma_j^T\right.\right)
\prod_i\mathcal N\left(\left(\begin{smallmatrix}z_i \\ \sigma_i\end{smallmatrix}\right)
\left| \, 0
, \frac{1}{N}\sum_\nu\left(\begin{smallmatrix}u_\nu \\ w_\nu \end{smallmatrix}\right)\left(\begin{smallmatrix}u_\nu \\ w_\nu \end{smallmatrix}\right)^T
\right.\right)\ . \nonumber
\end{align}
Here $\left(\begin{smallmatrix}z_i \\ \sigma_i\end{smallmatrix}\right)$ and $\left(\begin{smallmatrix}u_\nu \\ w_\nu \end{smallmatrix}\right)$ are vectors of size $n+1$. $\frac{1}{N}\sum_\nu u_\nu^2$ self-averages to $\rho_u:=\frac{1}{\alpha}\mathbb E_uu^2=\frac{1}{\alpha}$.
As for the CSBM the order parameters are
\begin{align}
& m_w^a = \frac{1}{N}\sum_\nu u_\nu w_\nu^a && m_\sigma^a=\frac{1}{N}\sum_i y_i\sigma_i^a \\
& Q_w^{ab} = \frac{1}{N}\sum_\nu w_\nu^aw_\nu^b && Q_\sigma^{ab} = \frac{1}{N}\sum_i\sigma_i^a\sigma_i^b
\end{align}
We introduce them via new $\delta$-Dirac functions:
\begin{align}
* \propto &\, \mathbb E_{u}\int \prod_{a,\nu}\mathrm dw_\nu^aP_W(w_\nu^a)\prod_i\mathrm dy_iP_o(y_i|z_i)\mathrm dz_i\prod_{a,i}\mathrm dh_i^a\mathrm d\sigma_i^a\prod_{a\le b}\mathrm d\hat Q_w^{ab}\mathrm dQ_w^{ab}\mathrm d\hat Q_\sigma^{ab}\mathrm dQ_\sigma^{ab}\prod_a\mathrm d\hat m_w^a\mathrm dm_w^a\mathrm d\hat m_\sigma^a\mathrm dm_\sigma^a \\
 & \quad\quad \prod_{a\le b}e^{\hat Q_w^{ab}\left(NQ_w^{ab}-\sum_\nu w_\nu^aw_\nu^b\right)+\hat Q_\sigma^{ab}\left(NQ_\sigma^{ab}-\sum_i\sigma_i^a\sigma_i^b\right)} \prod_ae^{\hat m_w^a(Nm_w^a-\sum_\nu u_\nu w_\nu^a)+\hat m_\sigma^a(Nm_\sigma^a-\sum_i y_i\sigma_i^a)} \nonumber \\
 & \quad\quad e^{-\beta t\sum_{a,i\in R}l(y_ih_i^a)-\beta t'\sum_{a,i\in R'}l(y_ih_i^a)}\prod_i N\left(h_i\left| c\sigma_i+\lambda y_im_\sigma, Q_\sigma\right.\right)
\mathcal N\left(\left(\begin{smallmatrix}z_i \\ \sigma_i\end{smallmatrix}\right)
\left| \, 0
, \left(\begin{smallmatrix}\rho_u & m_w^T \\ m_w & Q_w \end{smallmatrix}\right)
\right.\right) \nonumber \\
 = &\, \int \prod_{a\le b}\mathrm d\hat Q_w^{ab}\mathrm dQ_w^{ab}\mathrm d\hat Q_\sigma^{ab}\mathrm dQ_\sigma^{ab}\prod_a\mathrm d\hat m_w^a\mathrm dm_w^a\mathrm d\hat m_\sigma^a\mathrm dm_\sigma^a \prod_{a\le b}e^{N(\hat Q_w^{ab}Q_w^{ab}+\hat Q_\sigma^{ab}Q_\sigma^{ab})} \prod_ae^{N(\hat m_w^am_w^a+\hat m_\sigma^am_\sigma^a)} \\
 & \quad \left[\mathbb E_u\int \prod_a\mathrm dw^ae^{\psi_w^{(n)}(w)} \right]^{N/\alpha}
\left[\int \mathrm dyP_o(y|z)\mathrm dz\prod_a\mathrm dh^a\mathrm d\sigma^a e^{\psi_\mathrm{out}^{(n)}(h,\sigma;t)} \right]^{\rho N} \nonumber \\
 & \quad \left[\int \mathrm dyP_o(y|z)\mathrm dz\prod_a\mathrm dh^a\mathrm d\sigma^a e^{\psi_\mathrm{out}^{(n)}(h,\sigma;t')} \right]^{(1-\rho)N} ; \nonumber
\end{align}
where we defined
\begin{align}
\psi_w^{(n)}(w) &= \sum_a\log P_W(w^a)-\sum_{a\le b}\hat Q_w^{ab}w^aw^b-\sum_a\hat m_w^auw^a \\
 \psi_\mathrm{out}^{(n)}(h,\sigma;\bar t) &= -\beta \bar t\sum_al(yh^a)-\sum_{a\le b}\hat Q_\sigma^{ab}\sigma^a\sigma^b-\sum_a\hat m_\sigma^ay\sigma^a-\frac{1}{2}(h-c\sigma-\lambda ym_\sigma)^TQ_\sigma^{-1}(h-c\sigma-\lambda ym_\sigma) \nonumber \\
& \quad\quad{}-\frac{1}{2}\log\mathrm{det}\,Q_\sigma-\frac{1}{2}
\left(\begin{smallmatrix}z \\ \sigma\end{smallmatrix}\right)^T
\left(\begin{smallmatrix}\rho_u & m_w^T \\ m_w & Q_w \end{smallmatrix}\right)^{-1}
\left(\begin{smallmatrix}z \\ \sigma\end{smallmatrix}\right)
-\frac{1}{2}\log\mathrm{det}\left(\begin{smallmatrix}\rho_u & m_w^T \\ m_w & Q_w \end{smallmatrix}\right)\ .
\end{align}
We use the replica-symmetric ansatz: we set $\hat Q^{aa}=\frac{1}{2}\hat R$, $\hat Q^{ab}=-\hat Q$, $Q^{aa}=R$, $Q^{ab}=Q$, $\hat m^a=-\hat m$ and $m^a=m$. We define the variances $V=R-Q$ and $\hat V=\hat R+\hat Q$. We take the first order in $n$; and as before we have
\begin{align}
Q_\sigma^{-1} &= \frac{1}{V_\sigma}I_n-\frac{Q_\sigma}{V_\sigma^2}J_{n,n} + o(n) \\
\log\mathrm{det}\,Q_\sigma &= n\frac{Q_\sigma}{V_\sigma}+n\log(V_\sigma) + o(n)\ ;
\end{align}
we compute that
\begin{align}
\left(\begin{smallmatrix}\rho_u & m_w^T \\ m_w & Q_w \end{smallmatrix}\right)^{-1} &=
\left(\begin{smallmatrix}\frac{1}{\rho_u}+n\frac{m_w^2}{V_w\rho_u^2} & -\frac{m_w}{V_w\rho_u}(1,\ldots,1) \\
-\frac{m_w}{V_w\rho_u}(1,\ldots,1)^T & \frac{1}{V_w}I_n-\frac{1}{V_w^2}(Q_w-\frac{m_w^2}{\rho_u})J_{n,n} \end{smallmatrix}\right) \\
\log\mathrm{det}
\left(\begin{smallmatrix}\rho_u & m_w^T \\ m_w & Q_w \end{smallmatrix}\right)
 &= \log \rho_u+\frac{n}{V_w}(Q_w-\frac{m_w^2}{\rho_u})+n\log V_w + o(n)\ .
\end{align}
We can factorize the replica introducing scalar standard Gaussians:
\begin{align}
* \propto &\, \int \mathrm d\hat Q_w\mathrm d\hat V_w\mathrm dQ_w\mathrm dV_w\mathrm d\hat Q_\sigma\mathrm d\hat V_\sigma\mathrm dQ_\sigma\mathrm dV_\sigma\mathrm d\hat m_w\mathrm dm_w\mathrm d\hat m_\sigma\mathrm dm_\sigma e^{\frac{nN}{2}(\hat V_wV_w+\hat V_wQ_w-V_w\hat Q_w+\hat V_\sigma V_\sigma+\hat V_\sigma Q_\sigma-V_\sigma\hat Q_\sigma)} \nonumber \\
& \quad e^{-nN(\hat m_wm_w+\hat m_\sigma m_\sigma)}
\left[\mathbb E_{u,\xi}\left(\int \mathrm dw\,e^{\psi_w(w)}\right)^n \right]^{N/\alpha}
\left[\mathbb E_{\xi,\zeta,\chi}\int \mathrm dy\mathrm dz\,\psi_\mathrm{out}^*(y,z)\left(\int \mathrm dh\mathrm d\sigma e^{\psi_\mathrm{out}(h,\sigma;t)}\right)^n \right]^{\rho N} \nonumber \\
& \quad \left[\mathbb E_{\xi,\zeta,\chi}\int \mathrm dy\mathrm dz\,\psi_\mathrm{out}^*(y,z)\left(\int \mathrm dh\mathrm d\sigma e^{\psi_\mathrm{out}(h,\sigma;t)}\right)^n \right]^{(1-\rho)N} \ ;
\end{align}
with
\begin{align}
\psi_w(w) =& \log P_W(w)-\frac{1}{2}\hat V_ww^2+\left(\xi\sqrt{\hat Q_w}+u\hat m_w\right)w \\
 \psi_\mathrm{out}(h,\sigma;\bar t) =& -\beta \bar tl(yh)-\frac{1}{2}\hat V_\sigma\sigma^2+\left(\xi\sqrt{\hat Q_\sigma}+y\hat m_\sigma\right)\sigma \\
 &\quad{}+\log\mathcal N\left(h|c\sigma+\lambda ym_\sigma+\sqrt{Q_\sigma}\zeta, V_\sigma\right)+\log\mathcal N\left(\sigma\left|\rho_u^{-1}m_wz+\sqrt{(1-\eta_w)Q_w}\chi, V_w\right.\right) \nonumber \\
\psi_\mathrm{out}^*(y,z) =& P_o(y|z)\,\mathcal N(z|0,\rho_u)\ ,
\end{align}
where we defined $\eta_w=\frac{m_w^2}{\rho_u Q_w}$.
We take the limit $N\to\infty$ and $n\to 0$. The free entropy is then
\begin{align}
\phi &= \frac{1}{2}\left(\hat V_wV_w+\hat V_wQ_w-V_w\hat Q_w+\hat V_\sigma V_\sigma+\hat V_\sigma Q_\sigma-V_\sigma\hat Q_\sigma\right)-\hat m_wm_w-\hat m_\sigma m_\sigma
+\frac{1}{\alpha}\mathbb E_{u,\xi}\left(\log\int\mathrm dw\,e^{\psi_w(w)}\right) \nonumber \\
 &\quad{}+\rho\mathbb E_{\xi,\zeta,\chi}\left(\int\mathrm dy\mathrm dz\,\psi_\mathrm{out}^*(y,z)\log\int\mathrm dh\mathrm d\sigma e^{\psi_\mathrm{out}(h,\sigma;t)}\right)  \\
 &\quad{}+(1-\rho)\mathbb E_{\xi,\zeta,\chi}\left(\int\mathrm dy\mathrm dz\,\psi_\mathrm{out}^*(y,z)\log\int\mathrm dh\mathrm d\sigma e^{\psi_\mathrm{out}(h,\sigma;t')}\right) \ . \nonumber
\end{align}
As before we rescale the order parameters according to $\hat V\to\beta\hat V$, $\hat Q\to\beta^2\hat Q$, $\hat m\to\beta\hat m$ and $V\to\beta^{-1}V$ for both $w$ and $\sigma$, so in the limit $\beta\to\infty$ by Laplace's method the inner integrals are not degenerated. We define
\begin{align}
w^* = \argmax_w\psi_w(w) &\\
(h^*,\sigma^*) = \argmax_{h,\sigma}\psi_\mathrm{out}(h,\sigma;\bar t=1) &\quad\quad (h^{'*},\sigma^{'*}) = \argmax_{h,\sigma}\psi_\mathrm{out}(h,\sigma;\bar t=0)\ .
\end{align}

The fixed-point equations are
\begin{align}
& m_w = \frac{1}{\alpha}\mathbb E_{u,\xi}\,uw^* \quad && m_\sigma=\mathbb E_{\xi,\zeta,\chi}\int\mathrm dy\mathrm dz\,\psi_\mathrm{out}^*(y,z)y\left(\rho\sigma^*+(1-\rho)\sigma^{'*}\right) \\
& Q_w=\frac{1}{\alpha}\mathbb E_{u,\xi}(w^*)^2 \quad && Q_\sigma=\mathbb E_{\xi,\zeta,\chi}\int\mathrm dy\mathrm dz\,\psi_\mathrm{out}^*(y,z)\left(\rho(\sigma^*)^2+(1-\rho)(\sigma^{'*})^2\right) \\
& V_w=\frac{1}{\alpha}\frac{1}{\sqrt{\hat Q_w}}\mathbb E_{u,\xi}\,\xi w^* \quad && V_\sigma=\frac{1}{\sqrt{\hat Q_\sigma}}\mathbb E_{\xi,\zeta,\chi}\int\mathrm dy\mathrm dz\,\psi_\mathrm{out}^*(y,z)\xi\left(\rho\sigma^*+(1-\rho)\sigma^{'*}\right)
\end{align}
\begin{align}
& \hat m_w = \frac{1}{V_w}\mathbb E_{\xi,\zeta,\chi}\int\mathrm dy\mathrm dz\,\psi_\mathrm{out}^*(y,z)\left(\rho_u^{-1}z-\chi\frac{\rho_u^{-1}m_w}{\sqrt{(1-\eta_w)Q_w}}\right)(\rho\sigma^*+(1-\rho)\sigma^{'*}) \\
& \hat Q_w=\frac{1}{V_w^2}\mathbb E_{\xi,\zeta,\chi}\int\mathrm dy\mathrm dz\,\psi_\mathrm{out}^*(y,z)\left(\rho(\sigma^*-\rho_u^{-1}m_wz-\chi\sqrt{(1-\eta_w)Q_w})^2 \right. \\
& \quad\quad\quad \left. {}+(1-\rho)(\sigma^{'*}-\rho_u^{-1}m_wz-\chi\sqrt{(1-\eta_w)Q_w})^2\right) \nonumber \\
& \hat V_w=\frac{1}{V_w}\left(1-\frac{1}{\sqrt{(1-\eta_w)Q_w}}\mathbb E_{\xi,\zeta,\chi}\int\mathrm dy\mathrm dz\,\psi_\mathrm{out}^*(y,z)\chi\left(\rho\sigma^*+(1-\rho)\sigma^{'*}\right)\right) \\
& \hat m_\sigma = \frac{\lambda}{V_\sigma}\mathbb E_{\xi,\zeta,\chi}\int\mathrm dy\mathrm dz\,\psi_\mathrm{out}^*(y,z)y\left(\rho(h^*-c\sigma^*-\lambda ym_\sigma)+(1-\rho)(h^{'*}-c\sigma^{'*}-\lambda ym_\sigma)\right) \\
& \hat Q_\sigma=\frac{1}{V_\sigma^2}\mathbb E_{\xi,\zeta,\chi}\int\mathrm dy\mathrm dz\,\psi_\mathrm{out}^*(y,z)\left(\rho(h^*-c\sigma^*-\lambda ym_\sigma-\sqrt Q_\sigma\zeta)^2+(1-\rho)(h^{'*}-c\sigma^{'*}-\lambda ym_\sigma-\sqrt Q_\sigma\zeta)^2\right) \\
& \hat V_\sigma=\frac{1}{V_\sigma}\left(1-\frac{1}{\sqrt{Q_\sigma}}\mathbb E_{\xi,\zeta,\chi}\int\mathrm dy\mathrm dz\,\psi_\mathrm{out}^*(y,z)\zeta\left(\rho(h^*-c\sigma^*)+(1-\rho)(h^{'*}-c\sigma^{'*})\right)\right)
\end{align}

The average train and test losses can be computed by deriving $\phi$ wrt $t$ and $t'$ and taking it extremum by evaluating it at the fixed-point of these equations.

The integral on $z$ can be computed by the change of variable $\chi\to\frac{\chi}{\sqrt{1-\eta_w}}-\frac{\rho_u^{-1}m_wz}{\sqrt{(1-\eta_w)Q_w}}$. We obtain the expressions given in the main part, after simplification of the notations.

\subsection{Solution in the large regularization limit}
\label{sec:pointFixe-rGrand}
In this subsection we take $r\to\infty$; we state the solution to eqs.~\eqref{eq:répCsbmDébut}-\eqref{eq:répCsbmFin} and \eqref{eq:répGlmSbmDébut}-\eqref{eq:répGlmSbmFin} and we give the expression of the test accuracy of the GCN.

The following expressions can be derived considering $l(x)=(1-x)^2/2$ quadratic, without loss of generality, since at large regularization the weights $w$ and the output $h(w)$ of the GCN are small, and $l$ can expanded around 0 as a quadratic potential. As to the regularization $\gamma$ we take a $l_2$ regularization, as explained in the main part \ref{sec:consequences}.

\paragraph{CSBM}
The test accuracy of the GCN is
\begin{align}
\mathrm{Acc}_\mathrm{test} &= \frac{1}{2}\left(1+\mathrm{erf}\left(\frac{\lambda m_\sigma+cV_w\hat m_\sigma+c\sqrt\mu m_w}{\sqrt 2\sqrt{Q_\sigma+c^2V_w^2\hat Q_\sigma+c^2Q_w}}\right)\right)\ , \label{eq:prec-csbm}
\end{align}
the summary statistics being
\begin{align}
& m_w = \frac{\rho}{\alpha r}\sqrt\mu(\lambda+c) &&\quad\quad V_w = \frac{1}{\alpha r} &&\quad\quad Q_w = \frac{\rho}{\alpha r^2}\left(1+c^2(1-\rho)+\rho(1+\mu)(\lambda+c)^2\right) \\
& m_\sigma = \frac{\rho}{\alpha r}(1+\mu)(\lambda+c) &&\quad\quad V_\sigma = \frac{1}{\alpha r} &&\quad\quad Q_\sigma = \frac{\rho}{\alpha^2 r^2}\left((1+\alpha)(1+c^2(1-\rho)) \right. \\
& && && \quad\quad\quad\quad  \left. {}+\rho(1+\mu)(1+\mu+\alpha)(\lambda+c)^2\right) \nonumber \\
& \hat m_w = \rho\sqrt\mu(\lambda+c) && &&\quad\quad \hat Q_w = \rho+\rho(\lambda\rho+c)^2+(1-\rho)\lambda^2\rho^2 \\
& \hat m_\sigma = \lambda\rho && &&\quad\quad \hat Q_\sigma = \rho
\end{align}

\paragraph{GLM--SBM}
The test accuracy of the GCN is
\begin{align}
\mathrm{Acc}_\mathrm{test} &= \mathbb E_\chi \frac{1}{2} \left(1+\mathrm{erf}\left(\frac{1}{\sqrt 2}\chi\sqrt\frac{2\alpha}{\pi}\right)\right) \left(1+\mathrm{erf}\left(\frac{\lambda m_\sigma+cV_w\hat m_\sigma+c\sqrt Q_w\chi}{\sqrt 2\sqrt{Q_\sigma+c^2V_w^2\hat Q_\sigma}}\right)\right) \nonumber \\
 &= \int_{>0}\frac{\mathrm dz}{\sqrt{2\pi/\alpha}}e^{-\alpha z^2/2}\left(1+\mathrm{erf}\left(\frac{\lambda m_\sigma+cV_w\hat m_\sigma+cm_w\alpha z}{\sqrt 2\sqrt{Q_\sigma+c^2V_w^2\hat Q_\sigma+c^2(Q_w-\alpha m_w^2)}}\right)\right)\ , \label{eq:prec-glmSbm}
\end{align}
the summary statistics being
\begin{align}
& m_w = \frac{\rho}{\alpha r}\sqrt{\frac{2\alpha}{\pi}}(\lambda+c) &&\; V_w = \frac{1}{\alpha r} &&\quad Q_w = \frac{\rho}{\alpha r^2}\left(1+c^2(1-\rho)+\rho(1+2\alpha/\pi)(\lambda+c)^2\right) \\
& m_\sigma = \frac{\rho}{\alpha r}(1+2\alpha/\pi)(\lambda+c) &&\; V_\sigma = \frac{1}{\alpha r} &&\quad Q_\sigma = \frac{\rho}{\alpha^2 r^2}\left((1+\alpha)(1+c^2(1-\rho))\right. \\
& && && \quad\quad\quad \left.{}+\rho((1+2\alpha/\pi)(1+\alpha)+2\alpha/\pi)(\lambda+c)^2\right) \nonumber \\
& \hat m_w = \rho\sqrt{\frac{2\alpha}{\pi}}(\lambda+c) && &&\quad \hat Q_w = \rho+\rho(\lambda\rho+c)^2+(1-\rho)\lambda^2\rho^2  \\
& \hat m_\sigma = \lambda\rho && &&\quad \hat Q_\sigma = \rho
\end{align}

In the limit $\lambda\to\infty$ the maximizer $c^*$ of \eqref{eq:prec-glmSbm} is
\begin{align}
& c^* = \frac{1}{\lambda}\argmin_{\tilde c}e^{-2b\tau_\mathrm{GLM-SBM}^\infty + a^2\tau_\mathrm{GLM-SBM}^\infty}\left(1-\mathrm{erf}\left(\sqrt 2a\tau_\mathrm{GLM-SBM}^\infty\right)\right) \label{eq:cOpt-glmSbm} \\
& a = \sqrt\alpha\tilde c\frac{\sqrt{2\alpha/\pi}}{1+2\alpha/\pi}\ ,\quad\quad b = \frac{\tilde c}{1+2\alpha/\pi} - \frac{1}{2}\frac{\alpha\tilde c^2+(1+\alpha)/\rho}{(1+\alpha)(1+2\alpha/\pi)+2\alpha/\pi} \\
& \tau_\mathrm{GLM-SBM}^\infty = \frac{1+2\alpha/\pi}{2\left(1+\alpha+\frac{2\alpha/\pi}{1+2\alpha/\pi}\right)}
\end{align}

\section{Bayes-optimal performances}
\label{sec:eqBO}
In section \ref{sec:consequences} we compare the GCN to the Bayes-optimal performances. The Bayes-optimal performances on the CSBM and the GLM--SBM were derived by \cite{dz23csbm} and \cite{aPrioriGen19}. They can be expressed as a function of the fixed-point of a system of equations over three scalar quantities.

These works consider a non-directed SBM with symmetric adjacency matrix $A$ and symmetric fluctuations $\Xi$ in $A^\mathrm{g}=\frac{\lambda}{\sqrt N}yy^T+\Xi$. In our work for simplicity we take $\Xi$ non-symmetric. Then the corresponding $A$ and $A^\mathrm{g}$ can be mapped to a non-directed SBM by the transform $(A+A^T)/\sqrt 2$ and it is sufficient to rescale the snr $\lambda$ of the non-directed SBM by $\sqrt 2$ to have the same snr as for the directed SBM. So we set $\Delta_I=2\lambda^2$ the signal-to-noise ratio of the corresponding low-rank matrix factorization problem.

\subsection{CSBM}
The equations are given by \cite{dz23csbm} in its appendix. The self-consistent equations read
\begin{align}
& m^t = \frac{\mu}{\alpha}m_u^t+\Delta_Im_y^{t-1} \\
& m_y^t = \rho+(1-\rho)\mathbb E_{W}\left[\tanh\left(m^t+\sqrt{m^t}W\right)\right] \\
& m_u^{t+1} = \frac{\mu m_y^t}{1+\mu m_y^t}
\end{align}
where $W$ is a standard scalar Gaussian. Once a fixed-point $(m,m_y,m_u)$ is obtained the test accuracy is given by
\begin{equation}
\mathrm{Acc}_\mathrm{test} = \frac{1}{2}(1+\mathrm{erf}\sqrt{m/2})\ .\label{eq:precBOcsbm}
\end{equation}
In the large $\lambda$ limit we have $m_y\to 1$ and
\begin{equation}
\log(1-\mathrm{Acc}_\mathrm{test}) \underset{\lambda\to\infty}{\sim} -\lambda^2\ .
\end{equation}

\subsection{GLM--SBM}
The equations are given by \cite{aPrioriGen19}, only for the unsupervised case $\rho=0$. The supervised part can be inferred from the simpler case of Bayes-optimal inference on a GLM \cite{barbier2019optimal}. Then the supervised part and the unsupervised part are merged in a linear fashion as on the CSBM. We need the following (not normalized) density on $y$ and~$z$:
\begin{equation}
Q(y,z;B, A, \omega, V) = P_o(y|z)e^{-A/2+By}\frac{e^{-(z-\omega)^2/2V}}{\sqrt{2\pi V}}\ .
\end{equation}
We define the update functions
\begin{align}
& Z_\mathrm{out}(B, A, \omega, V) = \int\mathrm dy\mathrm dz\ Q(y,z;B, A, \omega, V)  && Z_\mathrm{out}^\mathrm{sup}(\omega, V) = \int\mathrm dz\ Q(+1,z;0, 0, \omega, V) \nonumber \\
& \quad\quad\quad = e^{-A/2}\left(\cosh B+\sinh(B)\mathrm{erf}(\omega/\sqrt{2V})\right)  && \quad\quad\quad = \frac{1}{2}\left(1+\mathrm{erf}(\omega/\sqrt{2V})\right) \\
& f_\mathrm{out} = \partial_\omega\log Z_\mathrm{out} && f_\mathrm{out}^\mathrm{sup} = \partial_\omega\log Z_\mathrm{out}^\mathrm{sup} \\
& f_y = \partial_B\log Z_\mathrm{out} &&
\end{align}
Then the self-consistent equations read
\begin{align}
& \hat m_u^t = \rho\mathbb E_{\eta}\left[Z_\mathrm{out}^\mathrm{sup}\left(\sqrt{m_u^t}\eta, \rho_u-m_u^t\right)f_\mathrm{out}^\mathrm{sup}\left(\sqrt{m_u^t}\eta, \rho_u-m_u^t\right)^2\right]  \\
& \quad\quad{}+ (1-\rho) \mathbb E_{\xi,\eta}\left[Z_\mathrm{out}\left(\sqrt{\Delta_Im_y^t}\xi, \Delta_Im_y^t, \sqrt{m_u^t}\eta, \rho_u-m_u^t\right)f_\mathrm{out}\left(\sqrt{\Delta_Im_y^t}\xi, \Delta_Im_y^t, \sqrt{m_u^t}\eta, \rho_u-m_u^t\right)^2\right] \nonumber \\
& m_y^{t+1} = \rho+(1-\rho)\mathbb E_{\xi,\eta}\left[Z_\mathrm{out}\left(\sqrt{\Delta_Im_y^t}\xi, \Delta_Im_y^t, \sqrt{m_u^t}\eta, \rho_u-m_u^t\right)f_y\left(\sqrt{\Delta_Im_y^t}\xi, \Delta_Im_y^t, \sqrt{m_u^t}\eta, \rho_u-m_u^t\right)^2\right] \\
& m_u^{t+1} = \frac{1}{\alpha}\frac{\hat m_u^t}{1+\hat m_u^t}
\end{align}
where $\xi$ and $\eta$ are standard scalar Gaussians and $\rho_u=\alpha^{-1}$. Once a fixed-point $(\hat m_u,m_y,m_u)$ is obtained the test accuracy is given by
\begin{align}
& \mathrm{Acc}_\mathrm{test} = \mathbb E_{\xi,\eta}\left[\int\mathrm dy\mathrm dz\ Q\left(y,z;\sqrt{\Delta_Im_y}\xi, \Delta_Im_y, \sqrt{m_u}\eta, \rho_u-m_u\right) \delta_{y=\sign f_y\left(\sqrt{\Delta_Im_y}\xi, \Delta_Im_y, \sqrt{m_u}\eta, \rho_u-m_u\right)} \right] \\
&\quad = \mathbb E_{\eta}\left[ \frac{1}{2}\left(1+\mathrm{erf}\left(\frac{\sqrt{m_u}\eta}{\sqrt{2(\rho_u-m_u)}}\right)\right)\left(1+\mathrm{erf}\left(\frac{\sqrt{\Delta_Im_y}}{\sqrt 2}+\frac{1}{\sqrt{2\Delta_Im_y}}\mathrm{arcth}\,\mathrm{erf}\left(\frac{\sqrt{m_u}\eta}{\sqrt{2(\rho_u-m_u)}}\right)\right)\right)\right]\,. \label{eq:precBOglmSbm}
\end{align}
In the large $\lambda$ limit we have $m_y\to 1$ and
\begin{equation}
\log(1-\mathrm{Acc}_\mathrm{test}) \underset{\lambda\to\infty}{\sim} -\lambda^2\ .
\end{equation}

\section{Fashion-SBM, a semi-realistic dataset}
\label{sec:fashionSBM}
In fig.~\ref{fig:cOptimal} we introduced fashion-SBM to show that our prediction $c^*\approx 1/\lambda$ seems to hold for a dataset more realistic than the CSBM or the GLM--SBM. In this section we detail how fashion-SBM is constructed.

Fashion-SBM is made by populating a SBM with attributes from fashion-MNIST \cite{fashionMNIST17}. The binary labels $y$ of the nodes are drawn first. The graph is generated according to the SBM described in the main part, with parameters $d$ and $\lambda$. As to the features, we consider only the training set of fashion-MNIST; out of the ten classes we keep only two classes to form $\tilde X\in\mathbb R^{N\times M}$ that is normalized according to
\begin{align}
& \hat X_{i\mu} = \tilde X_{i\mu}+\epsilon_{i\mu} \\
& X_{i\mu} = \sqrt N\frac{\hat X_{i\mu}-\frac{1}{N}\sum_j\hat X_{j\mu}}{\sqrt{\sum_j(\hat X_{j\mu}-\frac{1}{N}\sum_k\hat X_{k\mu})^2}}
\end{align}
$\epsilon$ is a small noise added to each pixel to avoid pixels that are always black. The resulting dataset has dimensions $N=12000$ and $M=784$.

In the experiment \ref{fig:cOptimal} we choose to use the two classes 2 (pullover) and 4 (coat). They are similar enough to keep balanced the signals of the features and the graph. The other classes are more dissimilar and cary a stronger signal, which results in the graph having little effect on the performance.

\section{Details on numerics}
\label{sec:numériques}
The systems \eqref{eq:répCsbmDébut}-\eqref{eq:répCsbmFin} and \eqref{eq:répGlmSbmDébut}-\eqref{eq:répGlmSbmFin} are solved by the iterating the twelve equations in parallel until convergence. About twenty iterations are necessary. The iterations are stable and no damping is necessary. The integral over $(\xi, \zeta, \chi)$ is evaluated by Monte-Carlo over $10^6$ points; we use the same samples over the iterations so they can exactly converge. For the quadratic and hinge losses the extremizer of the potential \eqref{eq:potentielOut} has an explicit solution; for the logistic loss we compute it by Newton's descent, a few steps are enough. The whole computation takes around one minute on a single CPU core with 5GB of memory.

For figures \ref{fig:vsLambdaGrand} and \ref{fig:vsLambdaGrandVsR} solving these two systems we were only able to reach misclassification errors $1-\mathrm{Acc}_\mathrm{test}$ of $10^{-6}$ because of numerical imprecision and the finite number of Monte-Carlo samples.

\section{Supplementary figures}
\label{sec:figuresSup}
\subsection{Optimal architecture}
\begin{figure*}[ht]
 \centering
 \includegraphics[width=0.9\linewidth]{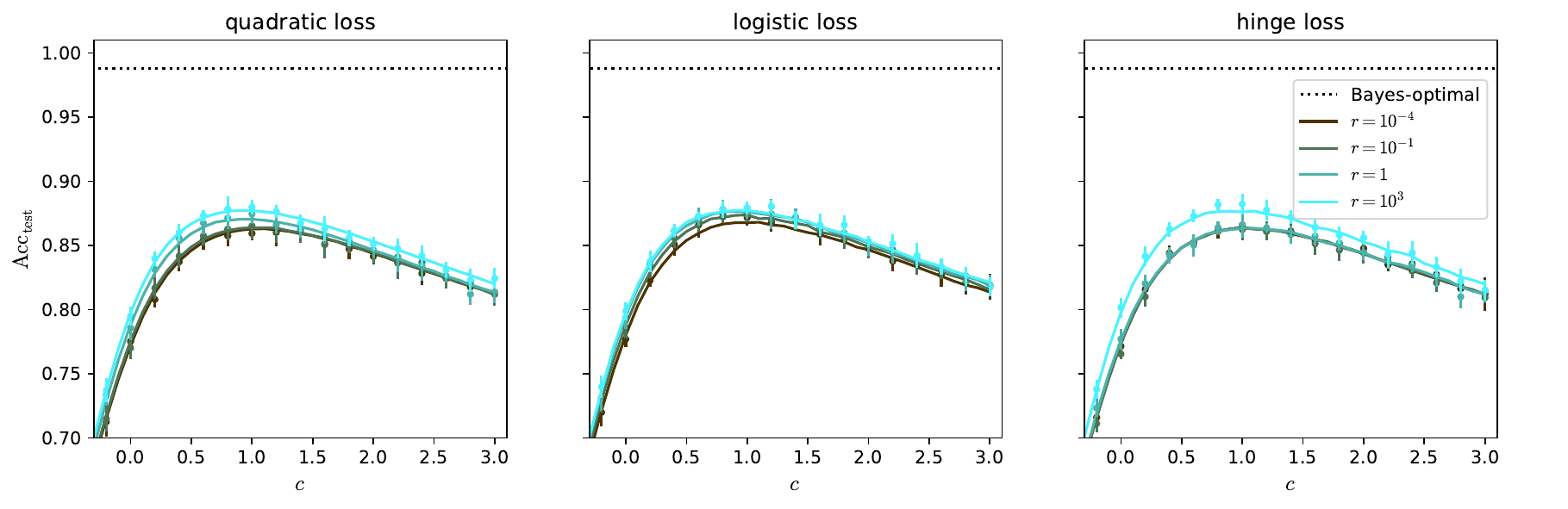}
 \includegraphics[width=0.9\linewidth]{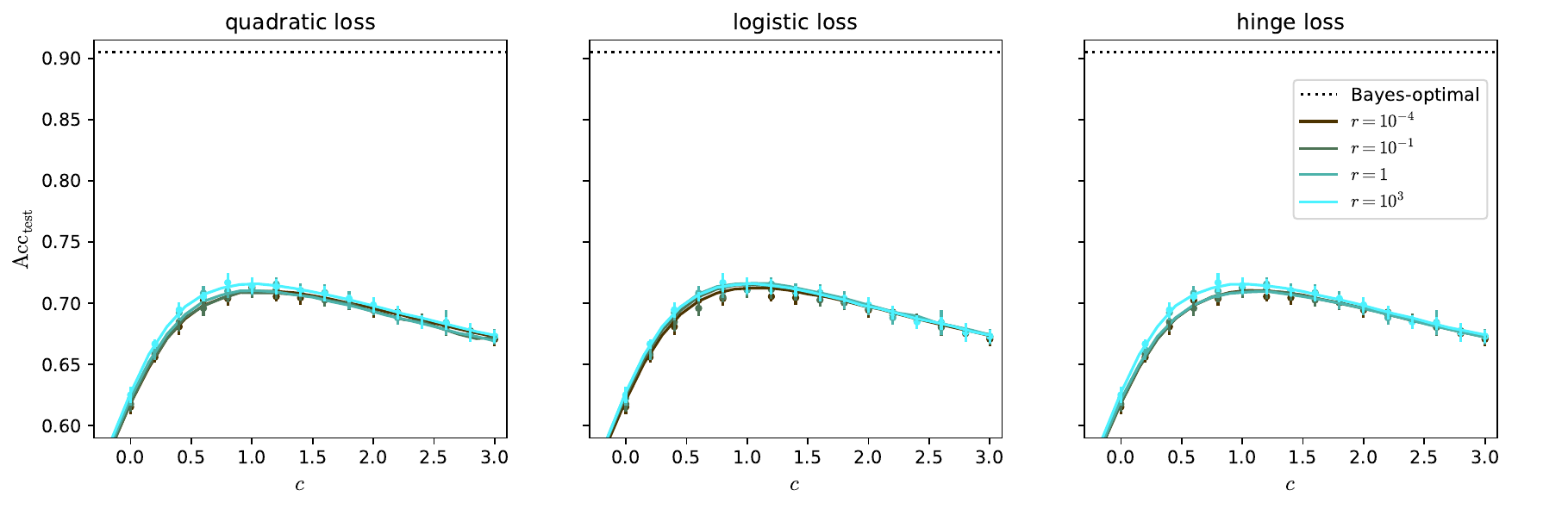}
 \caption{\label{fig:vsCR-alpha0.7} Search for the optimal parameters of the GCN. $\alpha=0.7$, $\rho=0.1$. \emph{Top:} CSBM, $\lambda=1.5$, $\mu=3$. \emph{Bottom:} GLM--SBM, $\lambda=1$. Full lines: prediction for the test accuracy obtained by eqs.~\eqref{eq:formuleAccErr}; dots: numerical simulation of the GCN for $N=10^4$ and $d=30$, averaged over ten experiments; dotted line: Bayes-optimal test accuracy.}
\end{figure*}

\begin{figure*}[t]
 \centering
 \includegraphics[width=0.9\linewidth]{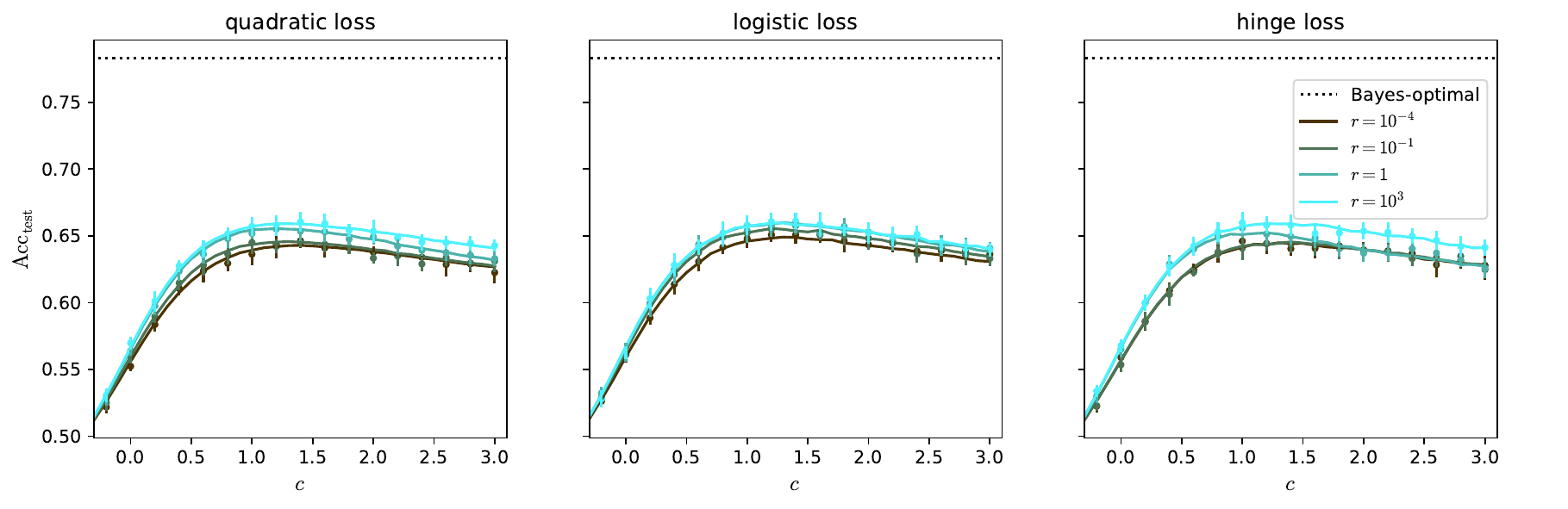}
 \includegraphics[width=0.9\linewidth]{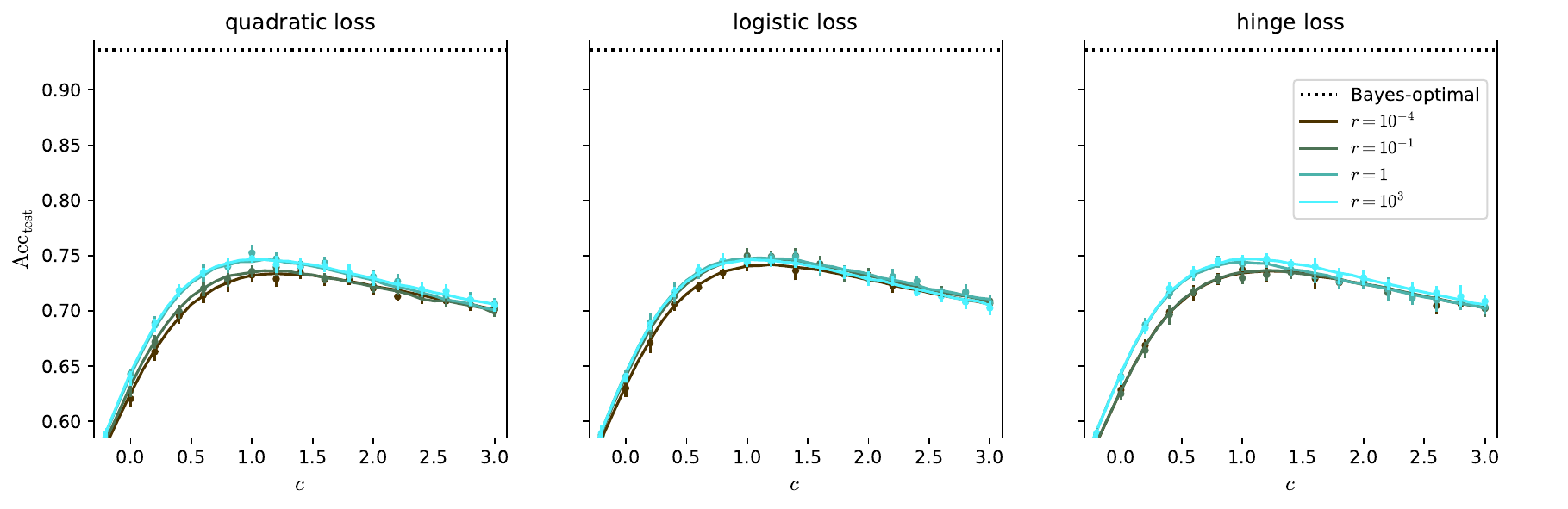}
 \caption{\label{fig:vsCR-alpha2} Search for the optimal parameters of the GCN. $\alpha=2$, $\rho=0.1$. \emph{Top:} CSBM, $\lambda=0.7$, $\mu=1$. \emph{Bottom:} GLM--SBM, $\lambda=1$. Full lines: prediction for the test accuracy obtained by eqs.~\eqref{eq:formuleAccErr}; dots: numerical simulation of the GCN for $N=10^4$ and $d=30$, averaged over ten experiments; dotted line: Bayes-optimal test accuracy.}
\end{figure*}

On figs.~\ref{fig:vsCR-alpha0.7} and \ref{fig:vsCR-alpha2} we search for the optimal architecture for data generated at different $\alpha$s, that is $\alpha=0.7$ and $\alpha=2$, for the CSBM and the GLM--SBM.
Together with figs.~\ref{fig:vsCR-csbm} and \ref{fig:vsCR-glmSbm} in the main part we reach conclusions that are detailed in section \ref{sec:perteReg}.

\newpage

\begin{figure}[h!t]
 \centering
 \includegraphics[width=0.6\linewidth]{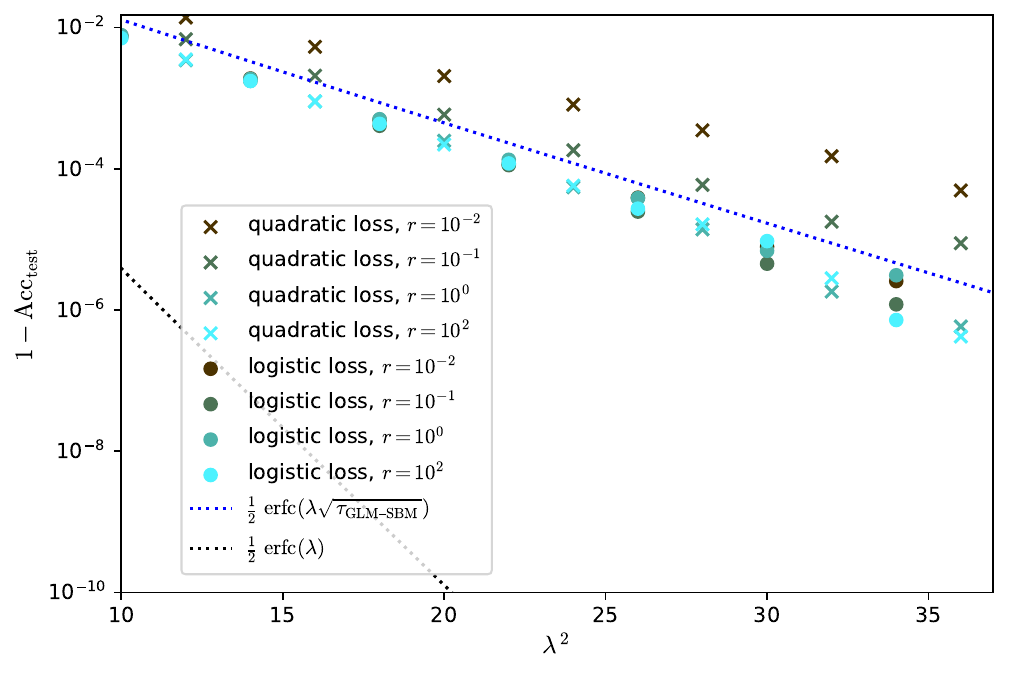}
 \caption{\label{fig:vsLambdaGrandVsR} Asymptotic misclassification error $1-\mathrm{Acc}_\mathrm{test}$ on the GLM--SBM. $\alpha=4$, $\rho=0.1$. Dots: prediction for the test accuracy obtained by eqs.~\eqref{eq:formuleAccErr} and \eqref{eq:répGlmSbmDébut}-\eqref{eq:répGlmSbmFin}, for $c=c^*$ optimal obtained by grid search. The blue dotted line is given by \eqref{eq:tauxGlmSbm}.}
\end{figure}

On fig.~\ref{fig:vsLambdaGrandVsR} we show the effect of the regularization $r$ on the convergence rate at large graph snr $\lambda$. For the quadratic loss, the rate depends on the regularization while for the logistic loss it does not.

\newpage

\subsection{Interpolation peak}
\begin{figure}[ht]
 \centering
 \includegraphics[width=0.9\linewidth]{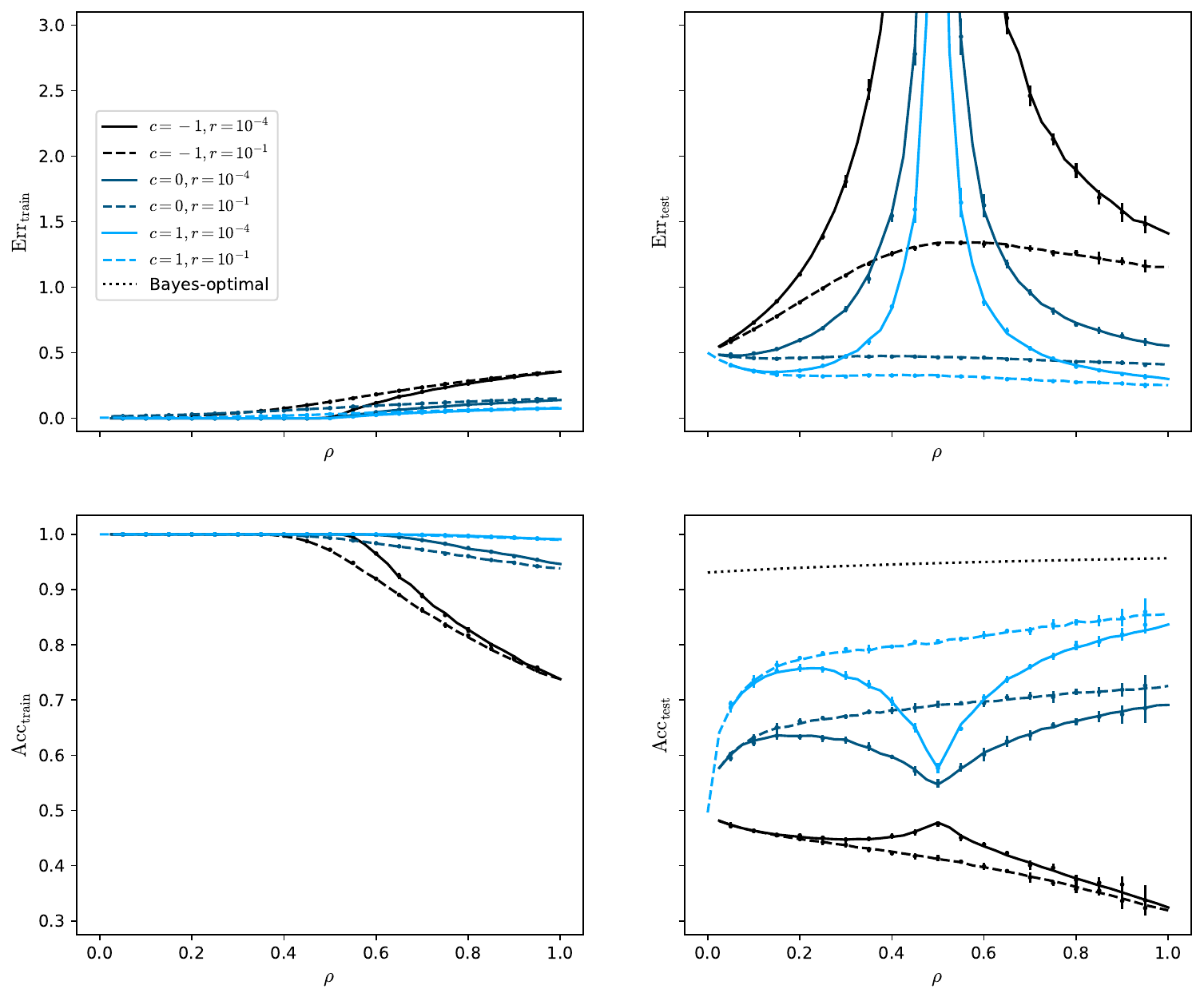}
 \caption{\label{fig:picsBis} Interpolation peak on the GLM--SBM for the quadratic loss. $\alpha=2$ and $\lambda=1$. Lines: predictions by eqs.~\eqref{eq:formuleAccErr} and \eqref{eq:répGlmSbmDébut}-\eqref{eq:répGlmSbmFin}; dots: numerical simulation of the GCN for $N=10^4$ and $d=N/2$, averaged over ten experiments; dotted line: Bayes-optimal test accuracy.}
\end{figure}

On fig.~\ref{fig:picsBis} we show that an interpolation peak appears for the ridge regression on the GLM--SBM when the regularization is small while varying the training ratio $\rho$. At the interpolation peak the train error becomes strictly positive, the train accuracy becomes strictly smaller than one, the test error diverges and the test accuracy has an inflexion point. The peak is located at $\alpha\rho=1$. Increasing the regularization $r$ smooths it out. Similar curves are obtained for the CSBM.

\begin{figure}[ht]
 \centering
 \includegraphics[width=0.9\linewidth]{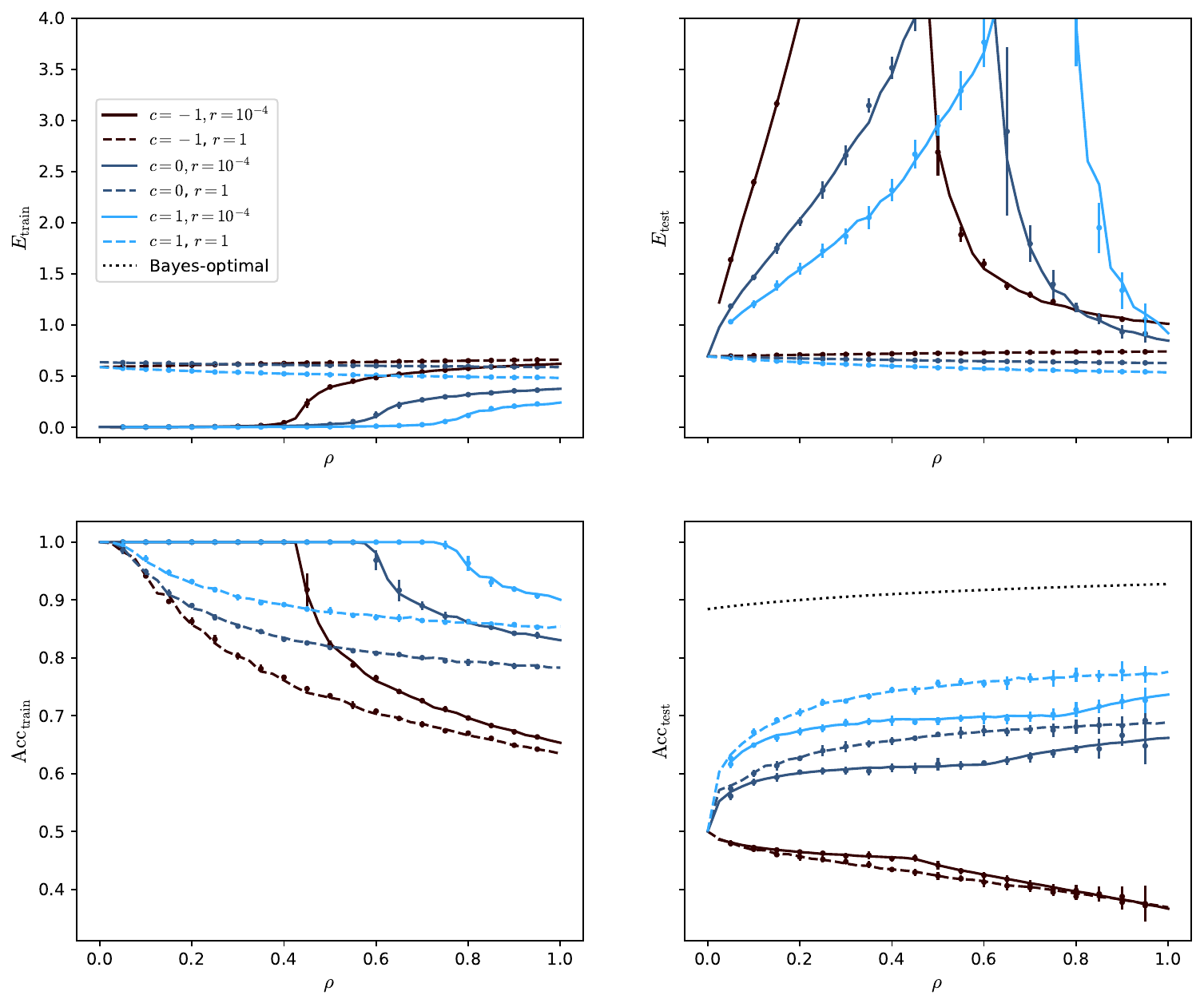}
 \caption{\label{fig:pics} Interpolation peak on the CSBM for the logistic loss. $\alpha=4$, $\lambda=1$ and $\mu=1$. Lines: predictions by eqs.~\eqref{eq:formuleAccErr} and \eqref{eq:répCsbmDébut}-\eqref{eq:répCsbmFin}; dots: numerical simulation of the GCN for $N=10^4$ and $d=N/2$, averaged over ten experiments; dotted line: Bayes-optimal test accuracy.}
\end{figure}

\begin{figure}[ht]
 \centering
 \includegraphics[width=0.6\linewidth]{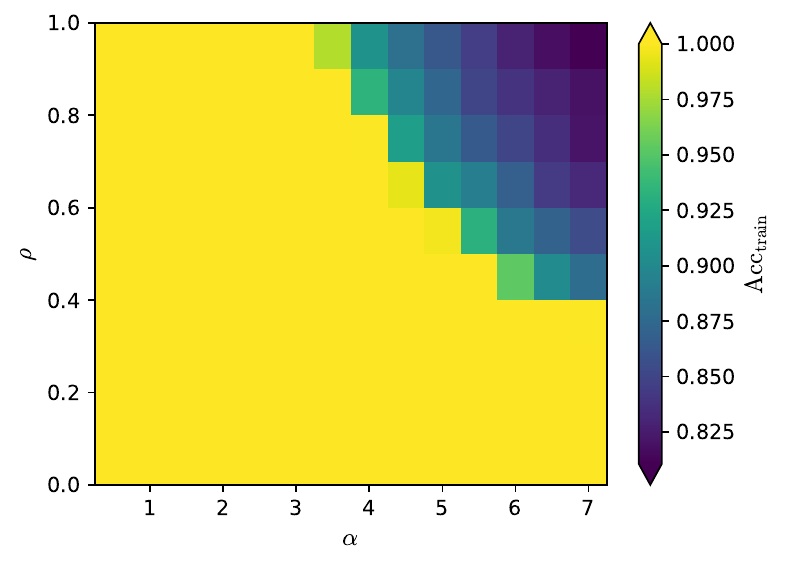}
 \caption{\label{fig:pics2d} Position of the interpolation peak on the CSBM for the logistic loss. The interpolation peak is located at the border of $\mathrm{Acc}_\mathrm{train}<1$. $\lambda=1$, $\mu=1$ and $c=1$. Predictions by eqs.~\eqref{eq:formuleAccErr} and \eqref{eq:répCsbmDébut}-\eqref{eq:répCsbmFin}.}
\end{figure}

On fig.~\ref{fig:pics} we show that an interpolation peak appears for the logistic regression on the CSBM when the regularization is small while varying the training ratio $\rho$. At the interpolation peak the train error becomes strictly positive, the train accuracy becomes strictly smaller than one, the test error diverges and the test accuracy has an inflexion point. The position of the peak depends on the self-loop intensity $c$ and the aspect ratio $\alpha$. On fig.~\ref{fig:pics2d} we show how its position varies with respect to $\rho$ and $\alpha$ at $c=1$. Increasing the regularization $r$ smooths it out. Similar curves are obtained for the hinge loss and the GLM--SBM.


\end{document}